\documentclass[pdflatex,sn-mathphys]{sn-jnl}% Math and Physical Sciences Reference Style
\jyear{2021}%

\theoremstyle{thmstyleone}%
\newtheorem{theorem}{Theorem}%  meant for continuous numbers
\usepackage{subfigure}
\usepackage{amsmath}
\usepackage{epstopdf}

\theoremstyle{thmstyletwo}%

\theoremstyle{thmstylethree}%
\newtheorem{definition}{Definition}%

\begin{document}

\title[GreenNet]{BI-GreenNet: Learning Green's functions by boundary integral network}

%% \equalcont{These authors contributed equally to this work.}
%%=============================================================%%

\author[1]{\fnm{Guochang} \sur{Lin}}\email{lingc19@tsinghua.edu.cn}

\author[2]{\fnm{Fukai} \sur{Chen}}\email{cfk19@mails.tsinghua.edu.cn}

\author[1,3]{\fnm{Pipi} \sur{Hu}}\email{hpp1681@gmail.com}

\author[4]{\fnm{Xiang} \sur{Chen}}\email{xiangchen.ai@huawei.com}

\author[2]{\fnm{Junqing} \sur{Chen}}\email{jqchen@tsinghua.edu.cn}

\author[5]{\fnm{Jun} \sur{Wang}}\email{jun.wang@cs.ucl.ac.uk}

\author*[1,3]{\fnm{Zuoqiang} \sur{Shi}}\email{zqshi@tsinghua.edu.cn}

%%% 我把 street 删掉了，下面是一个标准的
%%% \affil[3]{\orgdiv{Department}, \orgname{Organization}, \orgaddress{\street{Street}, \city{City}, \postcode{610101}, \state{State}, \country{Country}}}

\affil[1]{\orgdiv{Yau Mathematical Sciences Center}, \orgname{Tsinghua University}, \orgaddress{\city{Beijing}, \postcode{100084}, \state{Beijing}, \country{P.R. China}}}

\affil[2]{\orgdiv{Department of Mathematical Sciences}, \orgname{Organization}, \orgname{Tsinghua University}, \orgaddress{\city{Beijing}, \postcode{100084}, \state{Beijing}, \country{P.R. China}}}

\affil[3]{\orgname{Yanqi Lake Beijing Institute of Mathematical Sciences and Applications}, \orgaddress{\city{Beijing}, \postcode{101408}, \state{Beijing}, \country{P.R. China}}}

\affil[4]{\orgdiv{Noah’s Ark Lab}, \orgname{Huawei}, \orgaddress{\street{No. 3 Xinxi Road}, \city{Beijing}, \postcode{100085}, \state{Beijing}, \country{P.R. China}}}

\affil[5]{\orgname{University College London}, \orgaddress{\city{London}, \postcode{WC1E 6EA}, \state{London}, \country{United Kingdom}}}


\abstract{Green's function plays a significant role in both theoretical analysis and numerical computing of partial differential equations (PDEs). However, in most cases, Green's function is difficult to compute. The troubles arise in the following three folds. Firstly, compared with the original PDE, the dimension of Green's function is doubled, making it impossible to be handled by traditional mesh-based methods. Secondly, Green's function usually contains singularities which increase the difficulty to get a good approximation. Lastly, the computational domain may be very complex or even unbounded. 
To override these problems, we leverage the fundamental solution, boundary integral method and neural networks to develop a new method for
computing Green's function with high accuracy in this paper. We focus on Green's function of Poisson and Helmholtz equations in bounded domains, unbounded domains. We also consider Poisson equation and Helmholtz  domains with interfaces. Extensive numerical experiments illustrate the efficiency and the accuracy of our method for solving Green's function. In addition, we also use the Green's function calculated by our method to solve a class of PDE, and also obtain high-precision solutions, which shows the good generalization ability of our method on solving PDEs. }

\keywords{Green's function, Partial differential equation, Boundary integral, Neural network}


\pacs[MSC Classification]{35J08, 65N35, 65N80, 68T07}

\maketitle

\section{Introduction}\label{sec1}

Green's function plays a significant role in the theoretical research and engineering application of many widely-used partial differential equations (PDEs), such as the Poisson equation, Helmholtz equation, wave equation \cite{melnikov1977some}. For one thing, Green's function can help solve PDEs and develop PDE theory. Given the Green's function of the differential operator, the solutions of a class of PDE problems can be written explicitly in an integral form where the Green's function serves as the integral kernel \cite{duffy2015green, greenberg2015applications}. Thanks to this explicit representation of the PDE solution, Green's function provides a powerful tool to study the analytical properties of PDEs. For the other, Green's function also has applications in many physical and engineering fields such as quantum physics \cite{economou2006green}, electrodynamics \cite{jackson1999classical} and geophysics \cite{wapenaar2006green}.

Because of the importance of the Green's function, the computation of the Green's function, especially in general domains, has attracted more and more attention in the past several decades. Theoretically, in \citep{hancock2006method}, the analytical expression of Green's function of the Poisson equation of some simple cases is given, while \citep{kukla2012green} discussed Green's function of the Helmholtz equation interior or exterior of the unit disc. However, there is no analytical expression of the Green's function in general domains.

In the meanwhile, when its comes to computing Green's function numerically, there are also mainly three difficulties. Firstly, solving Green's functions is a high-dimensional problem. Compared with the original PDE, its dimension is doubled, which limits the application of traditional methods such as finite difference method (FDM) to solve Green's function directly. Secondly, Green's function is not smooth and has singular points, and thus more effort should be devoted to obtain high-precision solution. Thirdly, the domain of PDE may have complex geometrical structure or is even unbounded, adding more difficulty to the computation of Green's function.

Fortunately, the rapid development of neural network and deep learning in recent years \citep{goodfellow2016deep} open up new possibilities for computing the Green's function. Owing to its universal approximation ability \cite{hornik1989multilayer}, especially for high-dimensional functions, deep learning has made important progress in numerous fields such as image recognition \cite{krizhevsky2012imagenet}, natural language processing \cite{devlin2018bert} and many scientific computation problems. Note that the Green's function itself is the solution to a parametric PDE, we can consider using neural networks to solve this PDE.

In fact, tracing back to 1990s, there were works considering using neural networks to solve PDEs \cite{dissanayake1994neural,lagaris1998artificial}. In recent years, with the emergence of more powerful tools in machine learning, such as automatic differentiation \cite{baydin2018automatic}, more and more attention has been paid to this field. The most natural idea is to use neural networks to approximate the solutions of PDEs directly and use the residuals of the PDEs and the boundary conditions to construct loss functions for training, e.g. PINN \cite{raissi2019physics}, DGM \cite{sirignano2018dgm}. There are also many works that put forward different forms of loss functions, such as Deep Ritz\cite{weinan2018deep} using variational form of the PDE and MIM \cite{lyu2022mim} in which high-order PDEs are transformed to low-order systems. Except for the common choice of multilayer perceptron (MLP) as the network structure, in Deep Ritz method \cite{weinan2018deep}, the residual network (ResNet) structure is used while in \cite{cai2019multi} authors proposed multi-scale neural networks. In addition, there are also some works like Deep BSDE \cite{weinan2017deep} where PDEs are solved by combining stochastic differential equations and neural networks which can avoid extra automatic differentiation of the networks.

In addition to solving a single equation, solving parametric equations and learning solution operators, i.e., the mapping from the parameter or the source item of PDEs to the corresponding solution has drawn extensive attention most recently. In \cite{li2020fourier}, the Fourier neural operator (FNO) is proposed, where Fourier transformation is utilized to design the network architecture. In \cite{lu2019deeponet}, another network structure composed of branch net and trunk net dealing with the PDE parameters and spatial coordinates respectively is proposed. Deep Green \cite{gin2020deepgreen}, and MOD-net \cite{zhang2021mod} both use an analogous Green's function approximated by a neural network to represent the solution operator of nonlinear PDEs, which maps the source item or boundary value to the solution.

However, some important issues that are not fully considered in these existing neural network-based methods, together with the complexity of Green's function itself, hinder the direct application of these methods to solve Green's function. Firstly, most of the works learning PDE solution operators are based on supervised learning, which requires a large amount of accurate solution data as the supervisory signal. The data is often obtained by solving PDE repeatedly, making the computational cost of preparing the training data very high. Besides, the learned solution operator usually performs badly outside the training set, and thus the generalization ability of these methods is limited by the coverage of the dataset. Secondly, some works use the neural network to directly approximate Green's function. The singularity of Green's function is not taken into account, and extra differentiation with respect to the network input is required, which may degenerate the accuracy of the approximation. Lastly, for some problems such as electromagnetic wave propagation, solving PDEs in an unbounded domain is critical. However, existing methods use a neural network to approximate the solution operator or Green's function directly, which severely suffer from the difficulty of sampling in unbounded domains.

To address these issues and overcome the three difficulties in computing Green's function mentioned above, in this paper, we design a novel neural network-based formulation to compute Green's function directly. Firstly, we use the fundamental solution to decompose the Green's function into an explicit singular part and a smooth part such that the equation for Green's function is reformulated into a smooth high-dimensional equation. Neural network based methods are then designed to solve this high-dimensional PDE. In particular, we introduce two neural network formulation for this problem: derivative based GreenNet method (DB-GreenNet) and boundary integral equation based (BI-GreenNet) method. The idea of DB-GreenNet method is similar to PINN\citep{raissi2019physics}, DGM \citep{sirignano2018dgm} and some other articles \citep{berg2018unified}, which use the residual of equations and boundary conditions as the loss function, and directly approximate the objective function by a neural network. The derivative of the network with respect to network input appearing in the PDE residual is calculated by automatic differentiation. BI-GreenNet method is based on BINet \citep{lin2021binet}, which is a method for solving parameterized PDEs. Based on potential layer theory, the PDE solution is transformed into the a boundary integral equation such that the PDE is automatically satisfied and no extra differentiation with respect to network input is needed. As is verified by our experiments, BI-GreenNet method only outperforms the DB-GreenNet method, but also can compute the Green's function in unbounded domains.

Besides, the proposed neural network-based method can not only solve the Green's function of a single domain, but also that of interface problem. Interface problem refers to the problem in which an interface separates the computation domain into two parts, and the PDE parameters in the two parts are different, which widely appears in thermology \cite{friedman1975one}, fluid mechanics \cite{lee2003immersed}, electrodynamics \cite{jackson1999classical}, and many other fields. Some jump conditions across the interface is required such that the solution are usually discontinuous and non-smooth, making it a non-trival task to solve the interface problem. Therefore, solving the Green's function of the interface problem is also of great significance. Similarly, we can also derive the PDE for the Green's function of the interface problem and utilize the BI-GreenNet method to transform the PDE into the a boundary integral equation. Experiments also show that our method is of high accuracy.

In conclusion, the main advantages of our work are summarized as follows: First, we propose a neural network-based method to compute Green's function directly, which is a difficult problem for traditional methods to deal with. In addition, we also apply the Green's function computed by our formulation as the solution operator to solve a class of PDEs with high accuracy. Second, compared with other methods of learning solution operators of PDEs, exact solutions are not required as the data set in our formulation. It not only reduces the complexity of data preparation, but also improves the generalization ability of the calculated Green's function as the solution operator. Third, although the Green's function is very complex, we make full use of the properties of the Green's function itself, such that simple architecture of hidden layer network like MLP and ResNet is enough to fit the target function. Last, our formulation can not only compute the Green's function in any bounded domain, but also that of the interface problems. In particular, by utilizing the boundary integral equation, the BIE-based formulation can also solve the Green's function in unbounded domains.

%\section{Related Work}

%%% 补充到前面的讲算子学习的相关文献中去

%Because of the strong representation ability of neural networks, many works did not want to get only the mapping of spatial coordinates to solutions. They use the neural networks to find the mapping from parameter space to the solution space. The idea of the works \citep{li2020fourier,gin2020deepgreen} is closely related to Green's function. In addition, there is also work\cite{zhang2021mod} using the idea of the Green's function to calculate the solution operator. However, in the training process, these methods are difficult to avoid the dependence on the exact solution data and the information of the PDEs such as source term and boundary conditions.

The rest of this paper is organized as follows. In Section \ref{sec:Green}, we recall the Green's function and the basic theoretical foundation, potential theory of our method. In Section \ref{sec:method}, we introduce our BI-GreenNet method for solving Green's function. Extensive numerical experiments are shown in Section \ref{sec:num}. At last, conclusion remarks are made in Section \ref{sec:con}.

\section{Preliminaries}\label{sec:Green}
\subsection{Green's function}
In this paper, for simplicity, we focus on the Dirichlet Green's function, i.e., Dirichlet boundary condition is imposed on the boundary of the computation domain. Green's function of other types can be easily handled with some small modifications. We consider both problems defined on a single domain, which can be further divided into interior problem and exterior problem, and the interface problem defined on two domains with different PDE parameters separated by an interface, as will be elaborated below.

\begin{figure}[htbp]
\centering
\subfigure[Single domain]{
\centering
\includegraphics[width=0.4\textwidth]{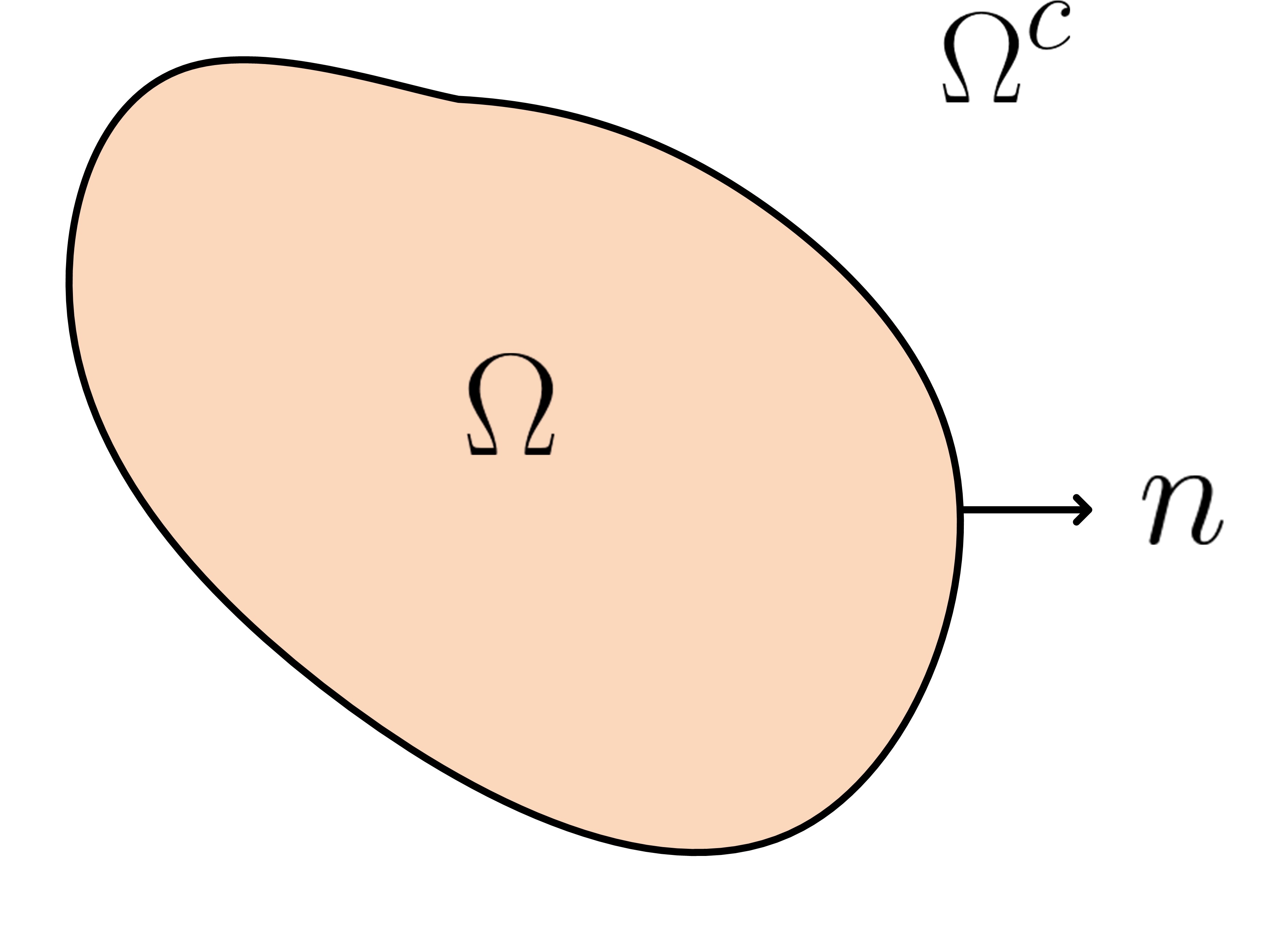}
\label{p1}
%\caption{fig1}
}%
\hspace{0.2in}
\subfigure[Interface problem]{
\centering
\includegraphics[width=0.4\textwidth]{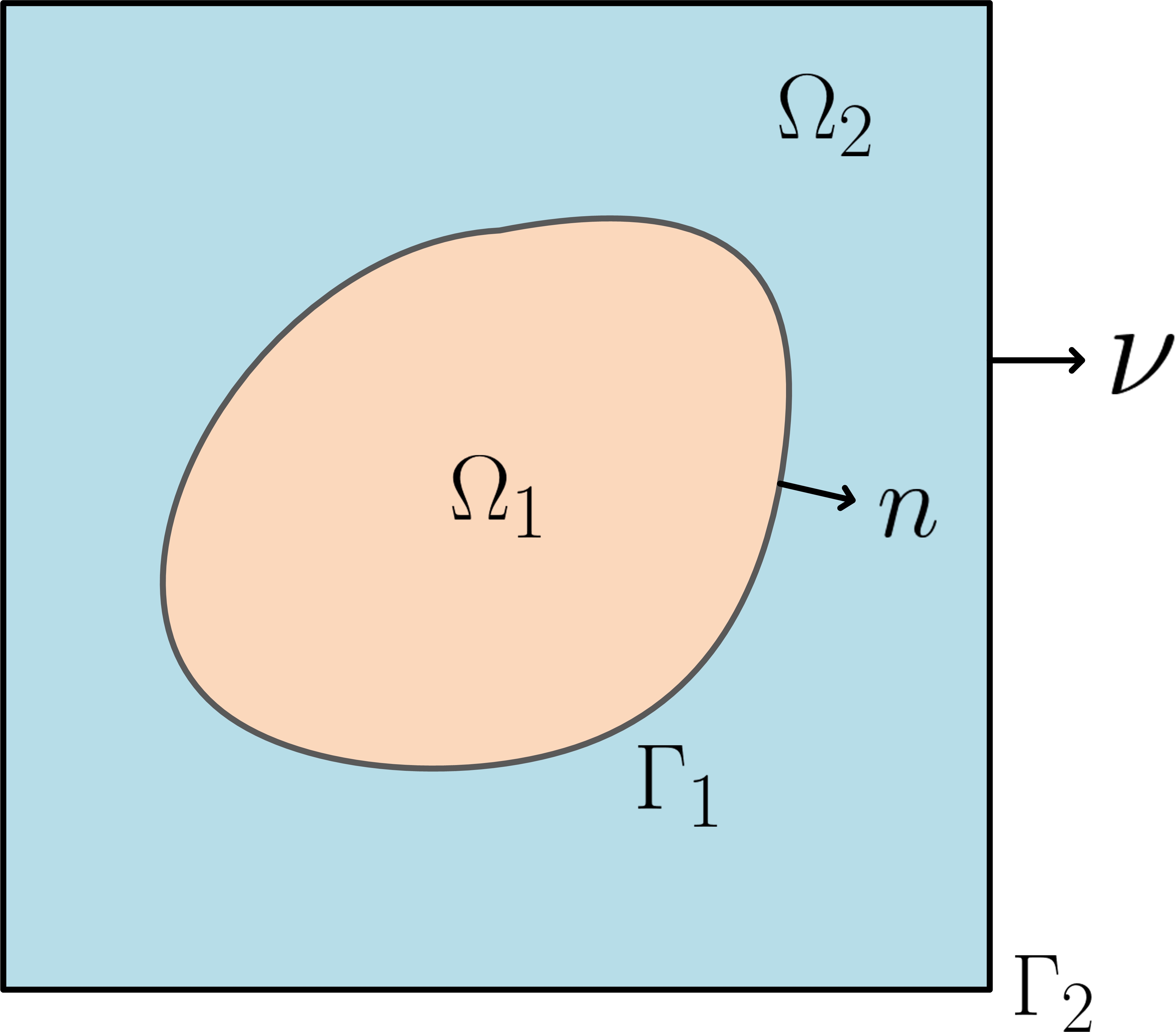}
\label{p2}}
\caption{Illustration of the computation domain}
\label{pint}
\end{figure}

\subsubsection{Green's function on a single domain}
 
As is presented in Fig. \ref{p1}, $\Omega\subset\mathbb{R}^d$ is a bounded domain, $\Omega^c=\mathbb{R}^d\backslash\Omega$. The interior problem and the exterior problem are formulated as
\begin{itemize}%[leftmargin=*]
    \item Interior problem,
    \begin{equation}
    \left\{
    \begin{aligned}
    \mathcal{L}u(x)=f(x) \ \text{in} \ \Omega,\\
    u(x) = g(x)  \ \text{on} \ \partial\Omega,
    \end{aligned}
    \label{sec3-01}
    \right.
    \end{equation}
    \item Exterior problem,
    \begin{equation}
    \left\{
    \begin{aligned}
    &\mathcal{L}u(x)=f(x) \ \text{in} \ \Omega^c,\\
    u(x) = g(x)  \ \text{on} \ \partial\Omega&+\ \text{some boundary conditions at infinity},
    \end{aligned}
    \label{sec3-02}
    \right.
    \end{equation}
\end{itemize}
where $\mathcal{L}$ is a differential operator. For brevity, we represent the equations in (\ref{sec3-01}) and (\ref{sec3-02}) as \begin{equation}
    \mathcal{L}u(x)=f(x)\label{sec3-0}\ \text{in}\ \Omega^*,
\end{equation} where $\Omega^*=\Omega$ or $\Omega^c$ for the interior or the exterior problem, respectively. 

In this paper, we focus on the Poisson equations and Helmholtz equations, i.e. $\mathcal{L}=-\Delta$ or $-\Delta-k^2$, where $k$ is the wave number. Helmholtz equations are the expansion of wave equation in frequency domain. It is used to describe wave propagation and is widely used in electromagnetics and acoustics. Because it involves wave propagation, Helmholtz equation often appears in the problem in unbounded domain, and because of the instability of Helmholtz equation, its numerical solution has always been an important problem.

For the equation (\ref{sec3-0}), we can use the corresponding Green's function $G(x)$ to give the analytical solution 
\begin{equation}
    u(x)=\int_{\Omega^*}G(x,y)f(y)dy+\int_{\partial\Omega}\frac{\partial G(x,y)}{\partial n_y}g(y)ds_y, 
\end{equation}
where $G(x,y)$ is a 2$d$-dimensional function satisfying 
\begin{equation}\label{G1}
\left\{
    \begin{aligned}
    \mathcal{L}_yG(x,y)=\delta(x-y), &\quad \forall \ x,y\in\Omega^{\ast},\\
    G(x,y)=0, & \quad \forall \ x\in\Omega^{\ast}, \ y\in\partial\Omega. 
    \end{aligned}
    \right.
\end{equation}

\subsubsection{Green's function of the interface problem}

As is shown in Fig. \ref{p2}, $\Omega\subset\mathbb{R}^d$ can be either bounded or unbounded. An interface $\Gamma_1\subset\mathbb{R}^{d-1}$ divides $\Omega$ into two regions, i.e., inside ($\Omega_1$) and outside ($\Omega_2)$ of the interface. The boundary of $\Omega$ is denoted as $\Gamma_2=\partial\Omega$.

The interface problem is then formulated as:

\begin{equation}\label{interface}
\left\{
\begin{aligned}
&\mathcal{L}u=f, &\quad &\text{in} \ \Omega,\\
&[u]=g_1, \
\left[\frac{1}{\mu}\frac{\partial u}{\partial n}\right]=g_2, &\quad &\text{on} \ \Gamma_1,\\
&u=g_3, &\quad &\text{on} \ \Gamma_2.
\end{aligned}
\right.
\end{equation}
where $\mathcal{L}$ is an operator with different parameters inside and outside the interface $\Gamma_1$, $n$
is the outward normal vector on the interface $\Gamma_1$. The bracket $[\cdot]$
denotes the jump discontinuity of the quantity approaching from $\Omega_2$ minus the one from $\Omega_1$. Moreover, some condition at infinity should be considered together for unbounded $\Omega$ to make the interface problem well-posed, which will be specified in detail in the experiments.

Similar to the single domain case, we focus on
\begin{itemize}
\item Poisson equations: $\mathcal{L}u=-\nabla\cdot(\frac{1}{\mu}\nabla u)$, where $\mu$ is a piecewise constant parameter such that $\mu=\mu_1$ in $\Omega_1$ and $\mu=\mu_2$ in $\Omega_2$;

\item Helmholtz equations: 
$\mathcal{L}u=-\nabla\cdot(\frac{1}{\mu}\nabla u)-\varepsilon k^2u$, where $\mu$ and $\epsilon$ are also piecewise constant parameters such that $\mu=\mu_1$, $\varepsilon=\varepsilon_1$ in $\Omega_1$ and $\mu=\mu_2$, $\varepsilon=\varepsilon_2$ in $\Omega_2$.
\end{itemize}

Utilizing the corresponding Green's function, the solution to the interface problem can be give by
\begin{equation}\label{sol_inter}
\begin{split}
u(x)&=\int_{\Omega}G(x,y)f(y)dy+\int_{\Gamma_1}\left(\frac{1}{\mu}\frac{\partial G(x,y)}{\partial n_y}g_1(y)-G(x,y)g_2(y)\right)ds_y\\
&\quad -\int_{\Gamma_2}\frac{1}{\mu}\frac{\partial G(x,y)}{\partial \nu_y}g_3(y)ds_y,
\end{split}
\end{equation}
where $\nu$ is the outward normal vector on $\Gamma_2$ and $G(x,y)$ satisfies
\begin{equation}\label{G_inter}
\left\{
\begin{aligned}
&\mathcal{L}_y G(x,y)=\delta(x,y), &\quad &\forall \ x,y\in\Omega,\\
&[G(x,y)]=\left[\frac{1}{\mu}\frac{\partial G(x,y)}{\partial n_y}\right]=0, &\quad &\forall \ x\in\Omega, \ y\in\Gamma_1,\\
&G(x,y)=0, &\quad &\forall \ x\in\Omega, \ y\in\Gamma_2.
\end{aligned}
\right.
\end{equation}

\subsection{Potential theory}
In this subsection, we briefly introduce the potential theory, the key to the boundary integral based (BIE-based) method, which will later be utilized to solve Green's function. We first define the single and double layer potential operators corresponding to the differential operator $\mathcal{L}_y$.

\begin{definition}\label{def.pt}
For any continuous function $h$ defined on $\Omega\times\partial\Omega$, the single layer potential is defined as
\begin{equation}
\mathcal{S}[h](x,y):=-\int_{\partial\Omega} G_0(y,z) h(x,z) d s_z, \label{Single}  
\end{equation}
and the double layer potential is defined as
\begin{equation}
\mathcal{D}[h](x,y):=-\int_{\partial\Omega} \frac{\partial G_0(y,z) }{\partial n_z} h(x,z) d s_z.
\label{Double}
\end{equation}
where $n_z$ denotes out normal of $\partial \Omega$ at $z$, and $G_0(x,y)$ is the fundamental solution corresponding to the differential operator $\mathcal{L}_y$ satisfying 
\begin{equation}
\mathcal{L}_yG_0(x,y)=\delta(x-y)\text{ for all }x,y\in\mathbb{R}^d.
\end{equation}
\end{definition}

Based on the layer potential theory of the Poisson equations and Helmholtz equations \cite{kellogg1953foundations}, we have the following theorem.
\begin{theorem} The single and double layer potentials have the following properties.

(i) The single layer potential $\mathcal{S}[h](x,y)$ and double layer potential $\mathcal{D}[h](x,y)$ are well defined in $\Omega^*\times\Omega^*$.

(ii) The single and double layer potentials satisfy equation \eqref{sec3-1} and (\ref{H_inter}), i.e.
\begin{equation}
\begin{aligned}
    \mathcal{L}_y\mathcal{S}[h](x,y)=0,\\ \mathcal{L}_y\mathcal{D}[h](x,y)=0.
\end{aligned}
\end{equation}
 
(iii) For $y_0\in\partial\Omega$, and the boundary near $y_0$ is smooth, we have
\begin{equation}%\label{thm:Property}
\begin{aligned}
\lim_{y\to y_0}\mathcal{S}[h](x,y) & = \mathcal{S}[h](x,y),\\
\lim_{y\to y_0^\pm}\mathcal{D}[h](x,y_0) & = \mathcal{D}[h](x,y_0)\mp\frac{1}{2}h(x,y_0),
\end{aligned}
\label{eq9:sec2}
\end{equation}
where $y\to y_0^-$ and $y\to y_0^+$ mean converging in $\Omega$ and $\Omega^c$ respectively.
\end{theorem}

By Theorem 1 we can see the solution to the PDE $\mathcal{L}_y u(x,y) =0$ can be written in a boundary integral form such that the PDE is satisfied automatically and we only need to solve the density function $h$ to fit the boundary condition.

\section{Neural network for Green's function }\label{sec:method}

\subsection{Removing the singularity }
As mentioned above, the Green's function is a singular function. However, we can change the equation (\ref{G1}) into a smooth equation by using the fundamental solutions $G_0$ of the original PDEs satisfying 
\begin{equation}
\mathcal{L}_yG_0(x,y)=\delta(x-y)\text{ for all }x,y\in\mathbb{R}^d,
\end{equation}
which also can be seen as the Green's function of the whole space $\mathbb{R}^d$. For many important PDEs, fundamental solutions can be written explicitly. For the Poisson equation $-\Delta u(x)=f(x)$ in $\mathbb{R}^2$, the fundamental solution is $G_0(x,y)=-\frac{1}{2\pi}\text{ln}\vert x-y\vert$, while the fundamental solution for the Helmholtz equation $-\Delta u(x)-k^2u(x)=f(x)$ in $\mathbb{R}^2$ is $G_0(x,y)=\frac{i}{4}H_0^1(k\vert x-y\vert)$ where $H^1_0$ is the Hankel function. The fundamental solutions of the high dimensional cases and more equations can be found in~\cite{hsiao2008boundary}. 

For the Green's function on a single domain, we set $H(x,y)=G(x,y)-G_0(x,y)$, then $H$ is a smooth function and satisfies the following equation
\begin{equation}
\left\{
    \begin{aligned}
    &\mathcal{L}_y H(x,y)=0, & \quad \forall \ x,y\in\Omega^{\ast},\\
    &H(x,y)=-G_0(x,y), & \quad \forall \ x\in\Omega^{\ast}, \ y\in\partial\Omega. 
    \end{aligned}
    \label{sec3-1}
    \right.
\end{equation}

For the Green's function of the interface problem, the fundamental solution of the Poisson equation in $\mathbb{R}^2$ is $G_0(x,y)=-\frac{\mu(y)}{2\pi}\ln\vert x-y\vert$, while the fundamental solution of the Helmholtz equation in $\mathbb{R}^2$ is $G_0(x,y)=\frac{i}{4}\mu(y)H_0^1(k\sqrt{\varepsilon(y)\mu(y)}\vert x-y\vert)$. Set $H(x,y)=G(x,y)-G_0(x,y)$, then $H$ satisfies the following equation
\begin{equation}\label{H_inter}
\left\{
\begin{aligned}
&\mathcal{L}_y H(x,y)=0, &\quad &\forall \ x,y\in\Omega,\\
&[H(x,y)]=-[G_0(x,y)], &\quad &\forall \ x\in\Omega, \ y\in\Gamma_1,\\
&\left[\frac{1}{\mu}\frac{\partial H(x,y)}{\partial n_y}\right]=-\left[\frac{1}{\mu}\frac{\partial G_0(x,y)}{\partial n_y}\right], &\quad &\forall \ x\in\Omega, \ y\in\Gamma_1,\\
&H(x,y)=-G_0(x,y), & \quad &\forall \ x\in\Omega, \ y\in\Gamma_2.
\end{aligned}
\right.
\end{equation}
After the singularity of the original equation is successfully eliminated, we will introduce how to use neural network to solve this problem by combining boundary integral method and neural network.

\subsection{BI-GreenNet}
After removing the singularity, we transform the problem (\ref{G1}) and (\ref{G_inter}) into problem (\ref{sec3-1}) and (\ref{H_inter}). Although the singularities have been removed, problem (\ref{sec3-1}) and (\ref{H_inter}) are still of high dimension such that traditional methods are difficult to handle. 
A natural idea to solve these problems is based on the automatic differentiation technique\cite{baydin2018automatic}. Like the Deep Ritz \cite{weinan2018deep} and PINN\citep{raissi2019physics}, for the interior problem in bounded domain, we use a neural network $\tilde{H}(x,y;\theta)$ to approximate the function $H(x,y)$ of the problems (\ref{sec3-1}) and (\ref{H_inter}). The derivative $\mathcal{L}_y\tilde{H}(x,y;\theta)$ can be calculated by automatic differentiation. The loss function can be designed as the following form by combining the the residuals of the PDE and the boundary condition
\begin{equation}
    \begin{aligned}
        L(\theta) = \sum_{i=1}^N \vert\mathcal{L}_y\tilde{H}(x_i^{\text{int}},y_i^{\text{int}};\theta)\vert^2+\lambda\sum_{j=1}^M\vert\tilde{H}(x_j^{\text{bd}},y_j^{\text{bd}};\theta)+G_0(x_j^{\text{bd}},y_j^{\text{bd}})\vert^2,
    \end{aligned}
\end{equation}
where $\lambda$ is the given weight and $\{x_i^{\text{int}},y_i^{\text{int}}\}_{i=1}^N$ are $N$ points randomly sampled in $\Omega\times \Omega$, while $\{x_i^{\text{bd}},y_i^{\text{bd}}\}_{j=1}^M$ are $M$ points randomly sampled in $\Omega\times \partial\Omega$. After training, an approximation of the Green's function can be obtained by the sum of $H(x,y)$ and the fundamental solution $G_0(x,y)$. We call this formulation derivative based GreenNet (DB-GreenNet) method. 

However, the derivative based method is not stable since extra differentiation with respect to the network input is needed, and the experiment results in Section \ref{sec:num} show the poor accuracy of DB-GreenNet. More importantly, it can not deal with the exterior problems since it is difficult to sample in unbounded domains.

\begin{figure}[htbp]
\centering
\includegraphics[width=0.9\textwidth]{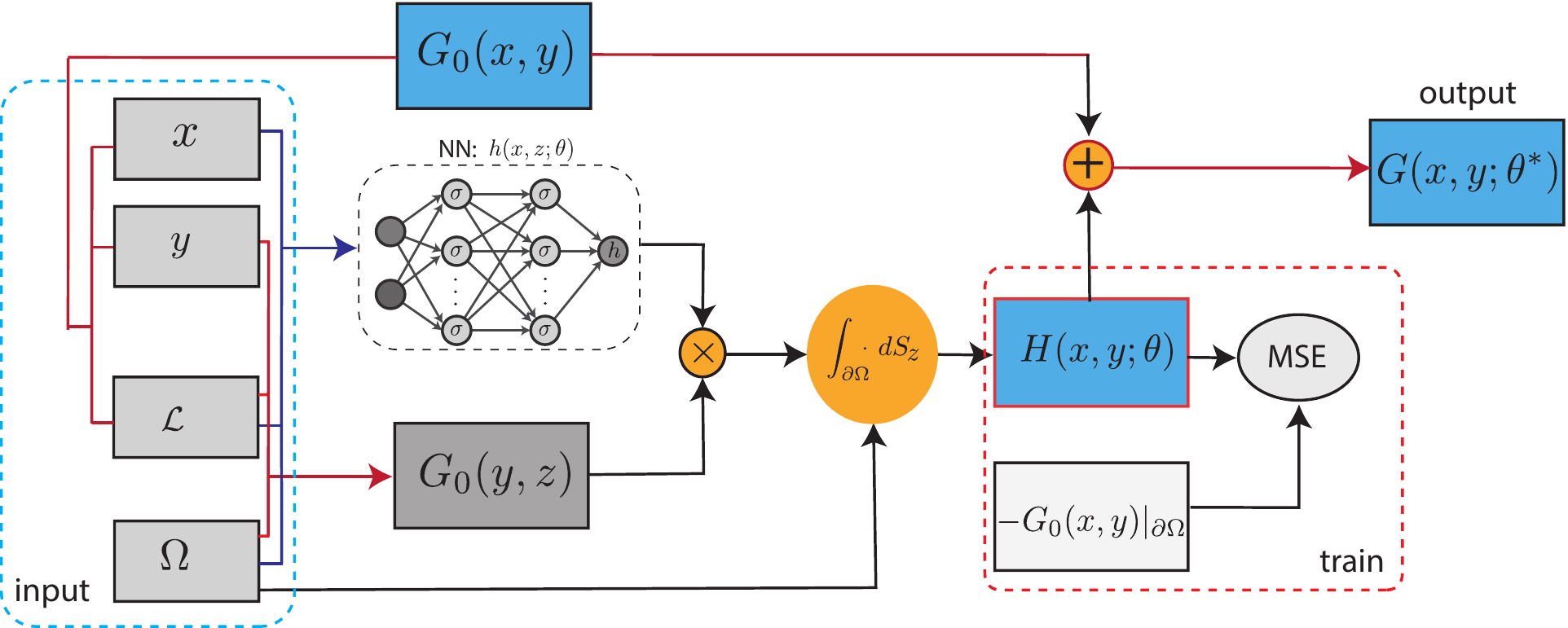}
\caption{The schematic of the boundary integral-based GreenNet (BI-GreenNet) method.}
\label{greennet}
\end{figure}

Therefore, we try to combine the Boundary Integral Network (BINet) \cite{lin2021binet}, a PDE solver that can achieve high accuracy and also can deal with the exterior problem in the unbounded domain, to compute the Green's function. Note that the partial differential equation $\mathcal{L}_y H(x,y)=0$ in both (\ref{sec3-1}) and (\ref{H_inter}) can be regarded as a parametric equation with $x$ being the parameter. As is illustrated in Fig. \ref{greennet}, based on BINet, we can represent the smooth component $H(x,y)$ in a boundary integral form such that the differential equation is satisfied automatically and only boundary conditions are to be fit. We call this fomulation boundary integral based GreenNet (BI-GreenNet) method and we will describe in detail how this method is implemented for both the single domain problem and the interface problem.

%For simplicity, we take problem (\ref{sec3-1}) as an example. The specific method for the interface problem can be obtained similarly.

%\subsubsection{Boundary Integral-Based method}

\textbf{Single domain problem.} For the single domain problems, because we mainly consider the Dirichlet problem in this paper, we choose the double layer potential operator expression in our formulation. $H(x,y)$ can then be written as the kernel integral form  
\begin{equation}
H(x,y)=\mathcal{D}[h](x,y):=-\int_{\partial\Omega} \frac{\partial G_0(y,z) }{\partial n_z} h(x,z) d s_z,
\label{eq5:sec3}
\end{equation}
where $G_0$ is the fundamental solution and the function $h(x,y)\in\Omega^*\times\partial\Omega$ is approximated by a hidden layer network, which can be selected as MLP or ResNet. The output of this network will satisfy the equation (\ref{sec3-1}) automatically, and we only need to fit the boundary condition, which the loss function is based on. This is also the reason that this method can handle the exterior PDE problem in unbounded domain. The loss function is 
\begin{equation}
\begin{aligned}
{L} = \sum_{i=1}^{N}\vert\mathcal{D}[h](x_i,y_i)\mp\frac{1}{2}h(x_i,y_i)+G_0(x_i,y_i)\vert^2,
\end{aligned}
\label{eq6:sec3}
\end{equation}
where $\{x_i,y_i\}_{j=1}^N$ are $N$ points randomly sampled in $\Omega\times \partial\Omega$. It can be seen that compared with the derivative-based method, the sampling points in the boundary integral method is much smaller. Another thing to note is the kernel function of the integral of the potential is singular, so we use the high accuracy quadrature rules in \cite{kapur1997high} and \cite{alpert1999hybrid} for smooth boundary, and for the boundary of the polygon domain, we use Simpson's quadrature rule directly. 

\textbf{Interface problem.} For the interface problem, we use the single layer potential on $\Gamma_1$ and the double layer potential on $\Gamma_2$. $H(x,y)$ can then be written as
\begin{equation}\label{HH}
H(x,y)=
\left\{
\begin{aligned}
&-\int_{\Gamma_1} G_0(y,z) h_1(x,z) d s_z, & \quad y\in \Omega_1,\\
&-\int_{\Gamma_1} G_0(y,z) h_2(x,z) d s_z-\int_{\Gamma_2} \frac{\partial G_0(y,z)}{\partial \nu_z} h_3(x,z) d s_z, & \quad y\in \Omega_2,
\end{aligned}
\right.
\end{equation}
where $h_1,h_2,h_3$ are approximated by three neural networks. Recall that in (\ref{H_inter}), three conditions on the boundary needs to be satisfied, i.e., two jump conditions on $\Gamma_1$ and one boundary condition on $\Gamma_2$. Therefore, the loss function of the interface problem is a weighted summation of three loss functions
\begin{equation}\label{LL}
    {L}=\lambda_{\text{jump1}} {L}_{\text{jump1}}+\lambda_{\text{jump2}} {L}_{\text{jump2}}+{L}_{\text{bd}},
\end{equation}
where $\lambda_{\text{jump1}}$ and $\lambda_{\text{jump2}}$ are the corresponding weights. Denote $\{x_i\}_{i=1}^{N_1}$ as the $N_1$ points randomly sampled in $\Omega$, $\{y_j\}_{j=1}^{N_2}$ as the $N_2$ points randomly sampled in $\Gamma_1$, and $\{w_k\}_{k=1}^{N_3}$ as the $N_3$ points randomly sampled in $\Gamma_2$. The first jump condition $[H(x,y)]=-[G_0(x,y)]$ gives the loss
\begin{equation}
\begin{aligned}
{L}_{\text{jump1}}=&\sum_{i=1}^{N_1}\sum_{j=1}^{N_2}\left\vert-\int_{\Gamma_1} G_0(y_j,z) h_2(x_i,z) d s_z-\int_{\Gamma_2} \frac{\partial G_0(y_j,z)}{\partial \nu_z} h_3(x_i,z) d s_z\right.\\
&\left.+G_0(x_i,y_j^+)+\int_{\Gamma_1} G_0(y_j,z) h_2(x_i,z) d s_z-G_0(x,y_j^-)\right\vert^2,
\end{aligned}
\end{equation}
where $y^+$ and $y^-$ denote the outside and inside of $\Gamma_1$, respectively. The second jump condition $\left[\frac{1}{\mu}\frac{\partial H(x,y)}{\partial n_y}\right]=-\left[\frac{1}{\mu}\frac{\partial G_0(x,y)}{\partial n_y}\right]$ gives the loss

\begin{equation}
\begin{aligned}
{L}_{\text{jump2}}=&\sum_{i=1}^{N_1}\sum_{j=1}^{N_2}\left\vert\frac{1}{\mu_2}\left[-\int_{\Gamma_1} \frac{\partial G_0(y_j,z)}{\partial n_y} h_2(x_i,z) d s_z+\frac{1}{2}h_2(x_i,y_j)\right.\right.\\
&\left.-\int_{\Gamma_2} \frac{\partial}{\partial n_y}\left(\frac{\partial G_0(y_j,z)}{\partial \nu_z}\right) h_3(x_i,z) d s_z+\frac{\partial G_0(x_i,y_j^+)}{\partial n_y}\right].\\
&\left.+\frac{1}{\mu_1}\left[\int_{\Gamma_1} \frac{\partial G_0(y_j,z)}{\partial n_y} h_2(x_i,z) d s_z+\frac{1}{2}h_1(x_i,y_j)-\frac{\partial G_0(x_i,y_j^-)}{\partial n_y}\right]\right\vert^2.
\end{aligned}
\end{equation}
The third loss is derived from the boundary condition $H(x,y)=-G_0(x,y)$

\begin{equation}
\begin{aligned}
{L}_{\text{bd}}=&\sum_{i=1}^{N_1}\sum_{k=1}^{N_3}\left\vert-\int_{\Gamma_1} G_0(w_k,z) h_2(x_i,z) d s_z-\int_{\Gamma_2} \frac{\partial G_0(w_k,z)}{\partial \nu_z} h_3(x_i,z) d s_z\right.\\
&\left.+\frac{1}{2}h_3(x_i,w_k)+G_0(x_i,w_k)\right\vert^2.
\end{aligned}
\end{equation}

\textbf{Remark:} In this paper, we mainly consider the problem with one interface. In fact, the proposed method can be easily generalized to problems with multiple interfaces by setting different density functions on both sides of each interface and the boundary of the whole computation domain.

In conclusion, we summarize the algorithms of the DB-GreenNet and BI-GreenNet for the single domain and interface problem in Algorithm \ref{DBG} and \ref{BIG}, respectively.
\begin{algorithm}
\caption{Derivative based GreenNet method (DB-GreenNet)}\label{DBG}
\begin{algorithmic}[1]
\Require $n_1,n_2,m,Epoch,\lambda,l,$ neural network $H(x,y;\theta)$ with parameters $\theta$ 
%\Ensure $G$ 
\State $i\Leftarrow 0$
\While {$i<Epoch$}
        \If{$i$ mod $m$=0}
            \State Randomly sample $n_1$ points $(x_{j_1},y_{j_1})\in\Omega\times\Omega$ 
            \State Randomly sample $n_2$ points $(x_{j_2},y_{j_2})\in\Omega\times\partial\Omega$
        \EndIf
        
        \State $L_1 \Leftarrow \sum_{j_1=1}^{n_1}\vert\mathcal{L}H(x_{j_1},y_{j_1};\theta)\vert^2$
        \State $L_2 \Leftarrow \sum_{j_2=1}^{n_2}\vert H(x_{j_2},y_{j_2};\theta)+G_0(x_{j_2},y_{j_2})\vert^2$
        \State $L \Leftarrow L_1 + \lambda L_2$
        \State $\theta \Leftarrow \theta - l\nabla_\theta L$
\EndWhile        
\State $G(x,y) \Leftarrow H(x,y;\theta)+G_0(x,y)$

\end{algorithmic}
\end{algorithm}

\begin{algorithm}
\caption{Boundary Integral based GreenNet method (BI-GreenNet)}\label{BIG}
\begin{algorithmic}[1]
\Require $n,m,Epoch,l,$ and the neural network $h(x,y;\theta)$ with parameters $\theta$ 
%\Ensure $G$ 
\State $i\Leftarrow 0$
\While {$i<Epoch$}
        \If{$i$ mod $m$=0}
            \State Randomly sample $n$ points $x_{j}\in\Omega^*$ 
        \EndIf
        \State Calculate loss function $L$ based on \eqref{eq6:sec3} for single domain problem and  \eqref{LL} for the interface problem
        \State $\theta \Leftarrow \theta - l\nabla_\theta L$
\EndWhile       
\State Calculate $H(x,y)$ based on \eqref{eq5:sec3} for single domain problem and \eqref{HH} for the interface problem
\State $G(x,y) \Leftarrow H(x,y)+G_0(x,y)$

\end{algorithmic}
\end{algorithm}

\section{Numerical Results}\label{sec:num}
In the implementation, we choose a ResNet structure neural network introduced in Deep Ritz method \citep{weinan2018deep}. In order to estimate the accuracy of the numerical solution $p$, we used the relative $L^2$ error $\Vert p-p^{\ast}\Vert_2/\Vert p^{\ast}\Vert_2$, where $p^*$ is the exact solution.
In the experiments, we choose the Adam optimizer to minimize the loss function, and we training the neural network on a single GPU of Tesla V100.  
\subsection{The Green's function of PDEs in bounded domains}\label{sec4-1}
In this subsection, we will use the GreenNet to compute Green's functions of Poisson equations and Helmholtz equations with different shapes of domains, then we will also use the Green's function solved by our methods to solve the PDEs with different source functions. 

\subsubsection{The Poisson Equation in the Unit Disc} 
In this experiment, we consider the Green's function of the Poisson equation in the unit disc. From the notes\citep{hancock2006method}, we can know the Green's function of the unit disc of the Poisson equation has the explicit expression,

\begin{equation}
    G(x,y)=\frac{1}{2 \pi} \ln \frac{r}{r^{\prime} \rho},
\end{equation}

where $y\in S^1$ and $y'=\frac{y}{\vert y\vert^2}$, and 
$r=\vert x-y\vert, \quad r^{\prime}=\vert x-y'\vert,\rho=\vert y\vert.$ Then we can compare the exact Green's function and the numerical Green's function calculated by GreenNet.

In the training process, for the BI-GreenNet method, we choose a ResNet with 8 blocks with 100 neurons per layer and ReLU activate function. For every 500 epochs, 80 new $x$ are randomly generated. After $5\times10^4$ epochs of training with learning rate $1\times10^{-5}$, we randomly generate 100 $x$ and 800 $y$ and compute the $G(x_i,y_j)$ of these $8\times10^4$ points $(x_i,y_j)$ to obtain an average relative $L_2$ error of the Green's function $G(x,y)$ of $4.66\times10^{-3}$. For the DB-GreenNet method, we used the neural network of same size. For every 500 epochs, $4\times10^4$ new $(x_i,y_i)\in\Omega\times\Omega$ and $1\times10^4$ new $(x_j,y_j)\in\Omega\times\partial\Omega$ are randomly generated for the PDE loss and boundary condition loss respectively. Similarly, after $5\times10^4$ epochs, the relative $L_2$ error of the Green's function $G(x,y)$ is $5.54\times10^{-2}$. The comparison of relative $L^2$ error can be seen in Table \ref{table-ex1}.

\begin{table}[htbp]
\caption{Relative $L^2$ error of the Green's function in section \ref{sec4-1}}
\label{table-ex1}
\begin{center}
\begin{tabular}{lcc}
\multicolumn{1}{c}{\bf structure}  &\multicolumn{1}{c}{\bf Poisson equation} &
\multicolumn{1}{c}{\bf Helmholtz equation} \\ \hline \\
BI-GreenNet  &7.44e-3 &8.56e-3 \\
DB-GreenNet  &5.54e-2 &8.75e-1 \\
\end{tabular}
\end{center}
\end{table}

\begin{figure}[t]
\centering
\subfigure[DB-GreenNet]{
%\label{confa}
\includegraphics[width=0.20\textwidth]{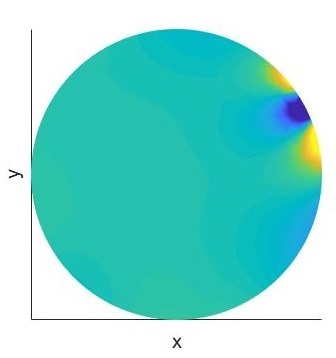}}
\centering
\subfigure[BI-GreenNet]{
%\label{confb}
\includegraphics[width=0.26\textwidth]{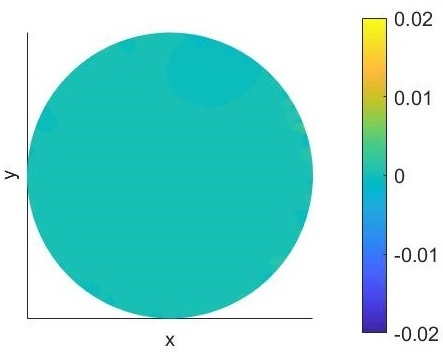}}
\centering
\subfigure[DB-GreenNet]{
%\label{confa}
\includegraphics[width=0.20\textwidth]{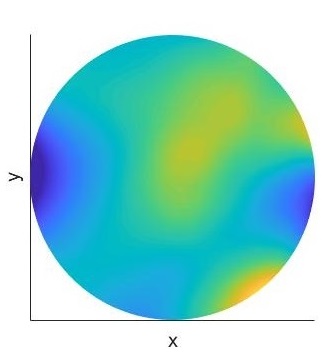}}
\centering
\subfigure[BI-GreenNet]{
%\label{confb}
\includegraphics[width=0.26\textwidth]{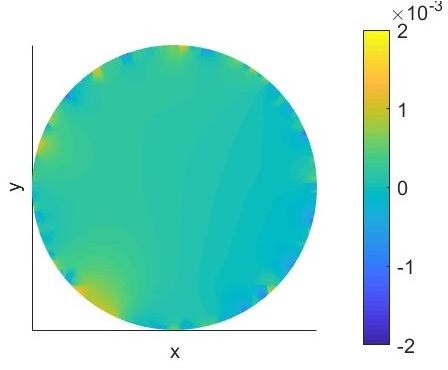}}

\caption{This figure show the difference between the exact Green's function $G(x,y)$ of the unit disc and Green's function calculated by DB-GreenNet and BI-GreenNet methods with fixed $x$. (a) and (b) are the results of $x=(0.8427,0.4386)$. (c) and (d) are the results of $x=(0.2923,0.0674)$.
}

\label{fig-ex1-1}
\end{figure}

The Fig. \ref{fig-ex1-1} show the difference of $y$ with fixed $x$ between the exact Green's function $G(x,y)$ of the unit disc and Green's function calculated by DB-GreenNet and BI-GreenNet methods. We can see the relative $L^2$ error of BI-GreenNet method still small when $x$ is close to the boundary. However, the error of derivative based method increases significantly.

\subsubsection{The Poisson Equation in the L-shaped Domain}
\begin{figure}[t]
\centering
\subfigure[source term]{
%\label{confa}
\includegraphics[width=0.3\textwidth]{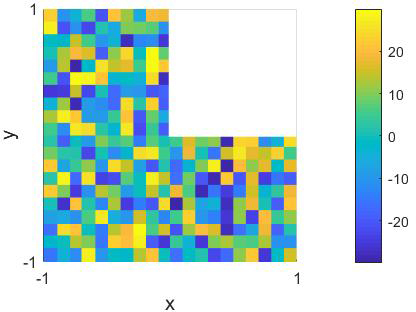}}
\hspace{0.03in}
\centering
\subfigure[exact]{
%\label{confb}
\includegraphics[width=0.3\textwidth]{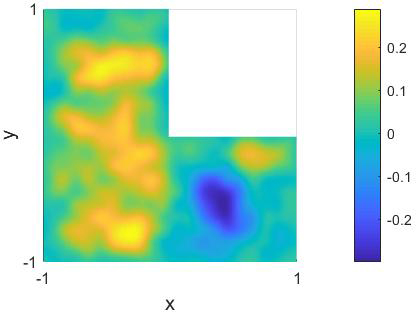}}
\hspace{0.03in}
\centering
\subfigure[numerical]{
%\label{confc}
\includegraphics[width=0.3\textwidth]{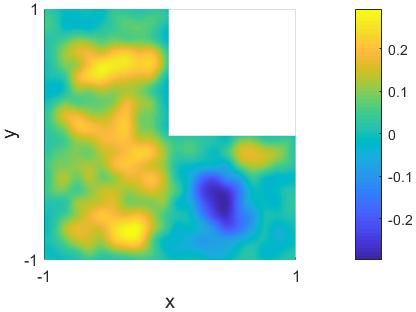}}

\subfigure[source term]{
%\label{confa}
\includegraphics[width=0.3\textwidth]{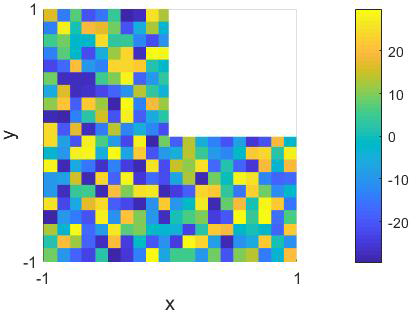}}
\hspace{0.03in}
\centering
\subfigure[exact]{
%\label{confb}
\includegraphics[width=0.3\textwidth]{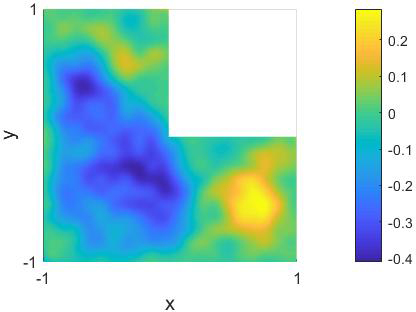}}
\hspace{0.03in}
\centering
\subfigure[numerical]{
%\label{confc}
\includegraphics[width=0.3\textwidth]{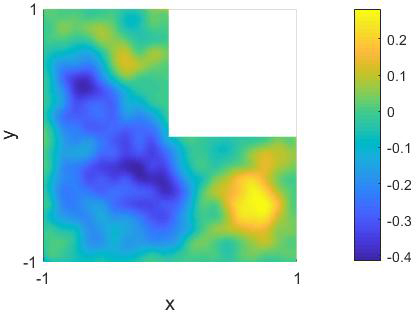}}

\centering
\caption{The exact solutions and numerical solutions of the problem \eqref{eq:4-1-2-1} with different source terms $f$. The (a) and (d) are two examples of the source functions and (b),(e) and (c),(f) are corresponding exact solutions and numerical solutions respectively.}
\label{fig:Ex1-2-1}
\end{figure}

In this experiment, we compute the Green's function of a L-shaped domain by BI-GreenNet method and use it to solve Poisson equations with different source functions in the same domain. We will compare with the results by finite difference method. We consider the following PDE problem,
\begin{equation}
    \begin{aligned}
    -\Delta u(x) = f(x) \text{ in }\Omega,\\
    u(x)=0\text{ on }\partial\Omega.
    \end{aligned}
    \label{eq:4-1-2-1}
\end{equation}
The $\Omega$ is a L-shaped domain and can be seen in the Fig. \ref{fig:Ex1-2-1}. We use a BINet with 6 blocks and 80 neurons per layer as the hidden layer network. After training, we also obtain the accurate approximation of Green's function $G(x,y)$ of the L-shaped domain, and then we use integral 
\begin{equation}
    u(x) = \int_\Omega G(x,y)f(y)dy
\end{equation}
to calculate the numerical solution of the PDE problem \eqref{eq:4-1-2-1}. In this experiment, we consider a piecewise constant source term $f$. Specifically, we divided $\Omega$ into 300 small rectangles and a constant in [-30,30] is randomly given on each small rectangle, respectively. The examples of the function $f$ can seen in the Fig. \ref{fig:Ex1-2-1} (a) and (d). The Fig. \ref{fig:Ex1-2-1} gives two example with randomly $f$, and shows that although the source function $f$ is complex, our method can also get accurate solutions. The exact
solutions are computed by finite difference method with the sufficiently small stepsize.

\begin{figure}[t]
\centering
%\label{confb}
\centering
\subfigure[relative $L^2$ error]{
%\label{confc}
\includegraphics[width=0.5\textwidth]{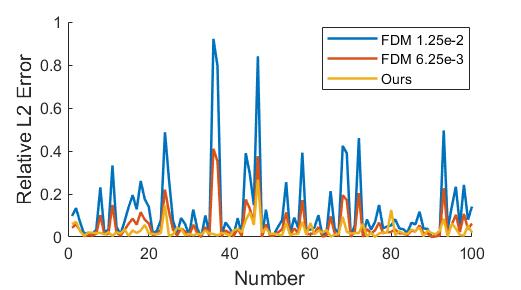}}
\centering
\subfigure[CDF of relative $L^2$ error]{
%\label{confc}
\includegraphics[width=0.4\textwidth]{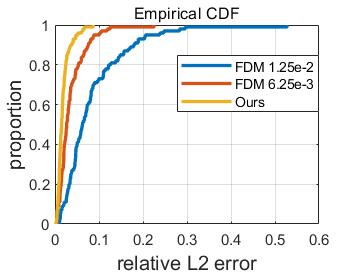}}

%\centering
%\subfigure[Time]{
%%\label{confc}
%\includegraphics[width=0.35\textwidth]{figure/ex1-2/Time_squ.jpg}}

\centering
\caption{The relative $L^2$ error of the solution of the problem \eqref{eq:4-1-2-1} with 100 different source terms $f$ under the finite different methods and BI-GreenNet method.}
\label{fig:Ex1-2-2}
\end{figure}

For comparison, we also used finite difference method (FDM) with two discrete stepsizes, 1/80 (=1.25e-2) and 1/160 (=6.25e-3) to solve this equation. The numbers of discrete points in $\Omega$ are 19521 and 77441 respectively. We randomly generate 100 different source functions $f$ and computed the relative $L^2$ errors with these methods, the results are shown in the Fig. \ref{fig:Ex1-2-2} and Table \ref{tab:1-2-1}. From these results, we can find that even if the number of parameters is less than the number of discrete points in the FDM method, our method can still achieve the higher accuracy, and it should be noted that the parameters of BI-GreenNet are used to represent the Green's function, the solution operator from the source function to the solution function, not just a single solution, but the value of the discrete points of FDM method only represent the single solution. It can be seen that BI-GreenNet represent more complex information with fewer parameters and achieve higher accuracy in this experiment. In the meanwhile, because using our method to solve the equations for different source term functions does not need to retrain the network, but only needs integration, so in the practical implementation, the speed of solving the equations is also very fast.

\begin{table}[t]
\caption{The average of relative $L^2$ error with 100 different source functions under the different methods of the problem \eqref{eq:4-1-2-1} and the number of parameters with these methods. }
\begin{center}
\begin{tabular}{lccc}
\multicolumn{1}{c}{\bf }  &\multicolumn{1}{c}{\bf FDM 1.25e-2} &
\multicolumn{1}{c}{\bf FDM 6.25e-3} &\multicolumn{1}{c}{\bf BI-GreenNet} \\ \hline \\
Average of relative $L^2$ error  & 0.1255 & 0.0559  & 0.0303 \\
Number of parameters  & 19521  & 77441   & 39281 \\
\end{tabular}
\end{center}
\label{tab:1-2-1}
\end{table}

\subsubsection{The Helmholtz Equation in the Square.}
In this experiment, we consider the Green's function of the following equation 
\begin{equation}
    \begin{aligned}
        -\Delta u(x) - 4 u(x)&=f(x)\text{ in }\Omega,\\
        u(x)&=0\text{ on }\partial\Omega,
    \end{aligned}
\end{equation}
where $\Omega=[-1,1]^2$. In this example, we will compare the performance of DB-GreenNet and BI-GreenNet on Helmholtz equation. The network structure and training details of the two methods are consistent with the first example. Because the Green's function of a Helmholtz equation is complex number, we add a output of the network to represent the imaginary part. In this example, for some fixed $x_i$, we can solve the equation (\ref{sec3-1}) about $y$ as the ground truth, but as mentioned before, it should be noted that the equation has to be solved again for each $x$, so the computation cost is very expensive. Table \ref{table-ex1} also show the relative $L^2$ error between ground truth and the numerical solution of Green's function. We can find DB-GreenNet method failed. This is because the oscillation of Helmholtz solution will increase the difficulty for derivative based method. The superiority of BI-GreenNet method over derivative based method is shown here. 

\begin{figure}[t]
\centering
\subfigure[distribution]{
\centering
\includegraphics[width=0.25\textwidth]{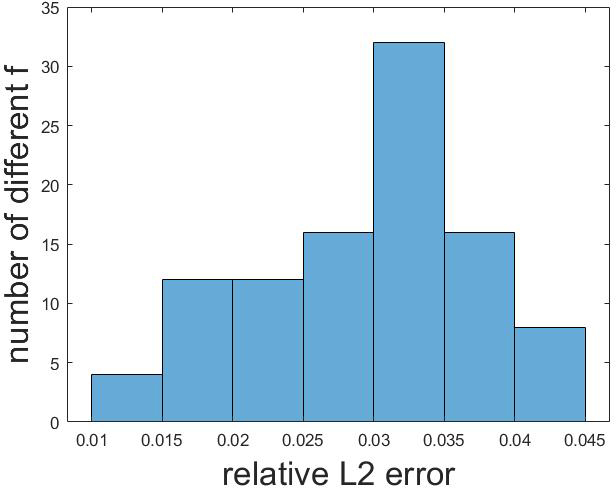}
%\caption{fig1}
}%
\subfigure[exact solution]{
\centering
\includegraphics[width=0.192\textwidth]{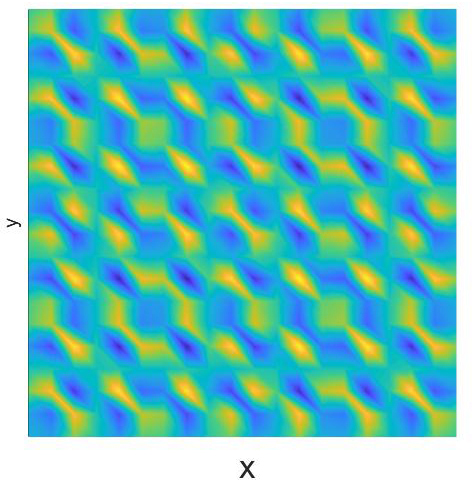}
%\caption{fig2}
}%
\subfigure[numerical solution]{
\centering
\includegraphics[width=0.248\textwidth]{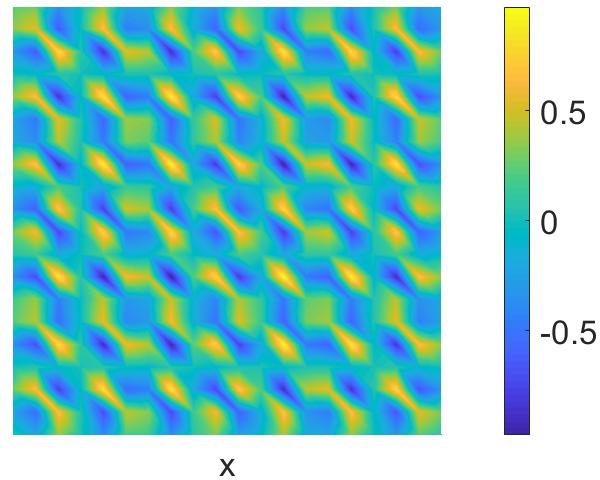}
%\caption{fig2}
}%
\subfigure[error]{
\centering
\includegraphics[width=0.25\textwidth]{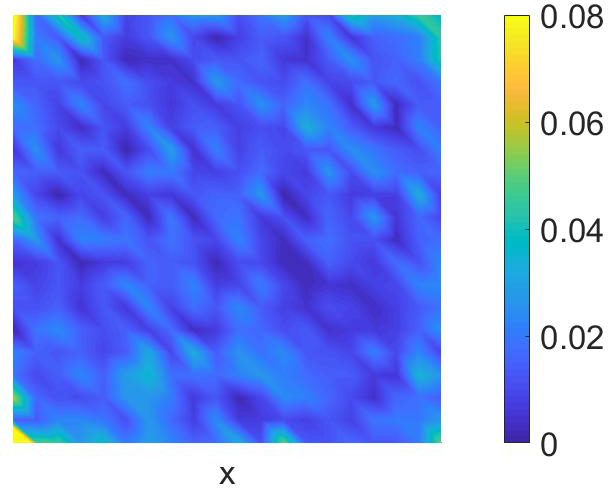}
%\caption{fig2}
}%
\caption{(a) is the distribution of relative $L^2$ error of the solution of the PDE with 100 different source terms; (b) and (c) is the real part of the exact solution and the numerical solution with the source term $f=(72\pi^2-4)\cos(6\pi x_1+\pi/2)\sin(6\pi x_2)$; (d) is the absolute error between exact solution and numercal solution.}
\label{fig-ex1-2}
\end{figure}

Next, we also show the ability on solving the Helmholtz equation in $\Omega$ by Green's function computed by BI-GreenNet method. We consider using the Green's function to solve the equations with different source terms $f$, where $f$ belongs to the following set
\begin{equation}
    \{((c_1^2+c_2^2)\pi^2-4)f_1(c_1\pi x_1)f_2(c_2\pi x_2): c_1,c_2=1,\cdots,5\},
\end{equation}
where $x=(x_1,x_2)\in\Omega$ and $f_1(a),f_2(a)\in\{\pm\sin(a)\}$. It can be seen that $f$ has 100 combinations. The histgram of the figure \ref{fig-ex1-2} show the distribution of the relative $L^2$ error between the exact solutions and the solutions calculated by our Green's function with different $f$. The relative $L^2$ error of the solution of the equation is between $1\times10^{-2}$ and $4.5\times10^{-2}$, which shows the stability of solving the equation with Green's function.

\subsection{The Green's function of PDEs in unbounded domains}
The exterior PDE problems in unbounded domains are common in scattering problems and electromagnetic problems, which is closely related to Helmholtz equation. Therefore, it will be very helpful for analyzing the properties of domains to study the Green's function of Helmholtz equation in unbounded domains. So in this subsection, we focus on the Green's functions of following Helmholtz equation problems
\begin{equation}
    \begin{aligned}
        -\Delta u(x) - k^2 u(x)&=f(x)\text{ in }\Omega^c,\\
        u(x)&=0\text{ on }\partial\Omega,\\
        \lim_{\vert x\vert\to\infty}(\frac{\partial}{\partial r}-ik)u(x)&=o(\vert x\vert^{-1/2}),
    \end{aligned}
    \label{ex2-0-1}
\end{equation}
where $k$ is the wavenumber and the limit of $\vert x\vert$ is called the Sommerfeld condition. In this subsection, we consider two shapes that often appear in practical problems, the bow-tie domain and the U-shaped domain.

We will compare the Green's function calculated by BI-GreenNet method and the exact Green's function. The exact Green's function $G(x,y)$ is derived by calculating the $H(x,y)$ of the problem \eqref{sec3-1} through boundary integral method first. However, as mentioned before, the equation \eqref{sec3-1} is a 4-dimensional problem, we can only solve this equation with some fixed points $x$. For the every given point $x$, it is required to solve the Helmholtz equation in an unbounded domain. Therefore, for a large number of different points $x$, the cost of obtaining $H(x,y)$ is very large. Although our method solves the Green's functions in unbounded domains, we cannot show the results in whole domains, and the domains concerned in practical problems is often bounded, so we will analyze the accuracy and show the results in sufficiently large bounded domains.  

\subsubsection{The Helmholtz equation out of the bow-tie antenna.} 
\begin{figure}[t]
\centering
\subfigure[exact,real]{
\centering
\includegraphics[width=0.180\textwidth]{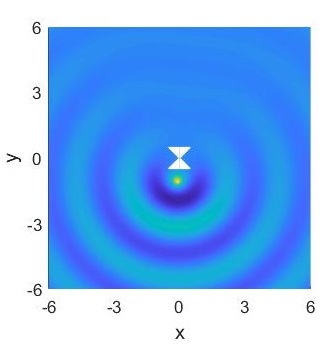}
%\caption{fig1}
}%
\subfigure[numerical,real]{
\centering
\includegraphics[width=0.194\textwidth]{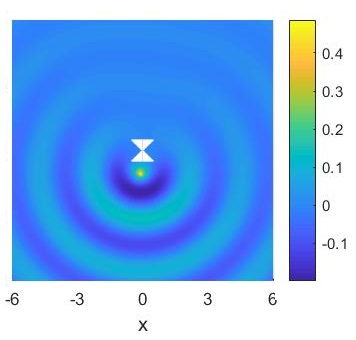}
%\caption{fig2}
}%
\subfigure[exact,imag]{
\centering
\includegraphics[width=0.178\textwidth]{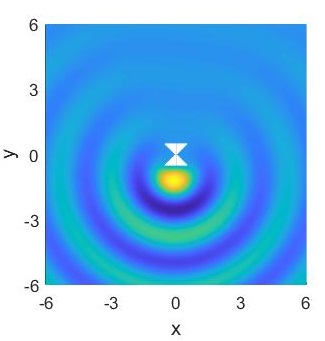}
%\caption{fig2}
}%
\subfigure[numerical,imag]{
\centering
\includegraphics[width=0.195\textwidth]{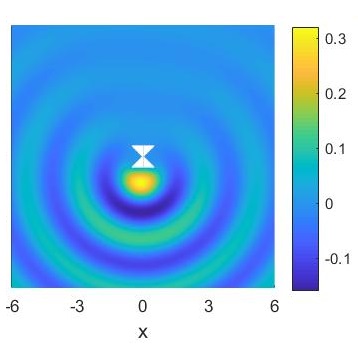}
%\caption{fig2}
}%
\subfigure[absolute error]{
\centering
\includegraphics[width=0.20\textwidth]{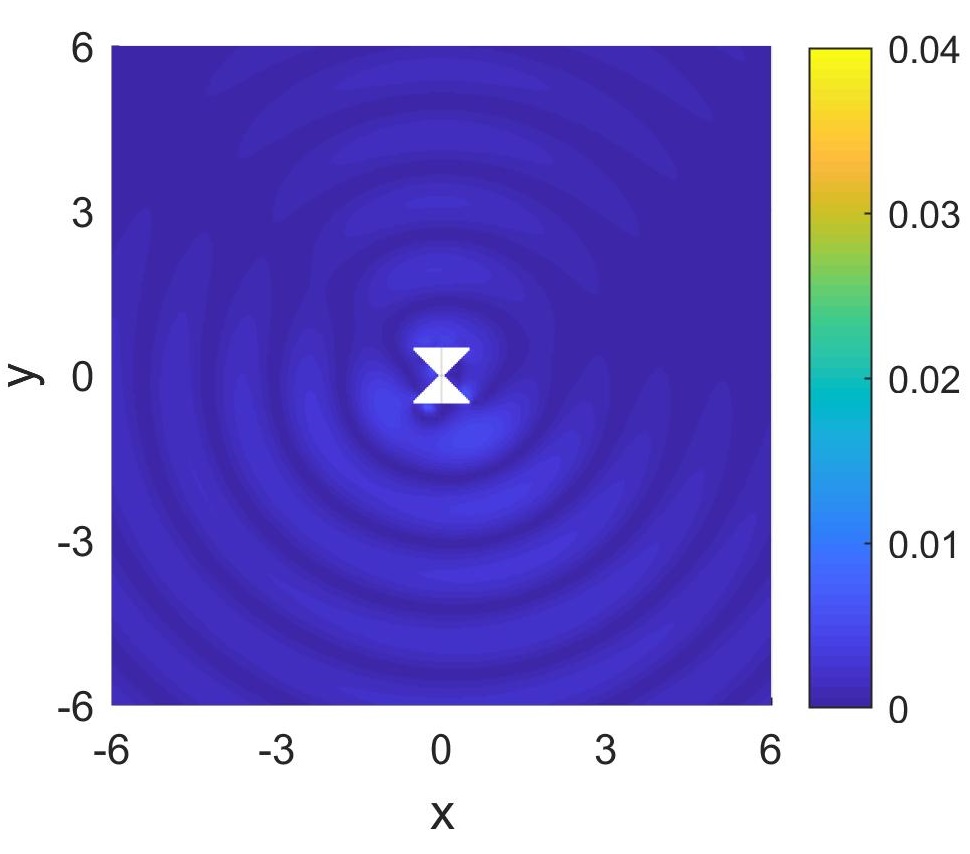}
%\caption{fig2}
}%

\centering
\subfigure[exact,real]{
\centering
\includegraphics[width=0.180\textwidth]{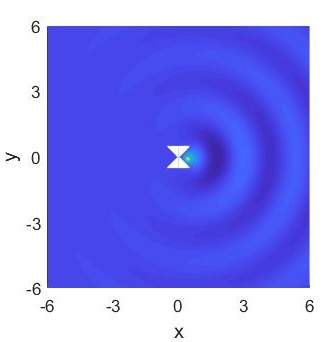}
%\caption{fig1}
}%
\subfigure[numerical,real]{
\centering
\includegraphics[width=0.194\textwidth]{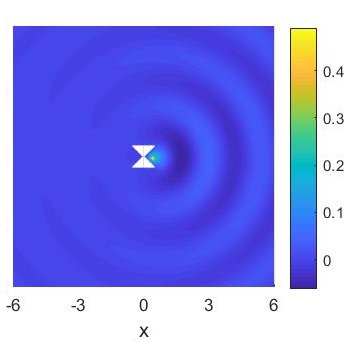}
%\caption{fig2}
}%
\subfigure[exact,imag]{
\centering
\includegraphics[width=0.178\textwidth]{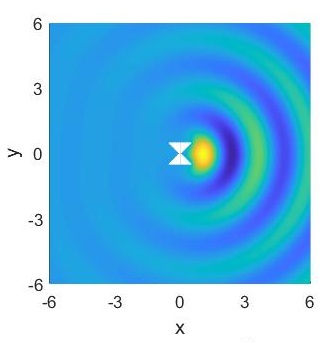}
%\caption{fig2}
}%
\subfigure[numerical,imag]{
\centering
\includegraphics[width=0.195\textwidth]{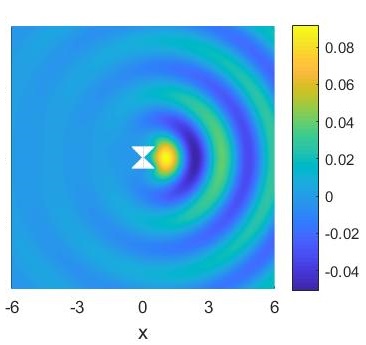}
%\caption{fig2}
}%
\subfigure[absolute error]{
\centering
\includegraphics[width=0.20\textwidth]{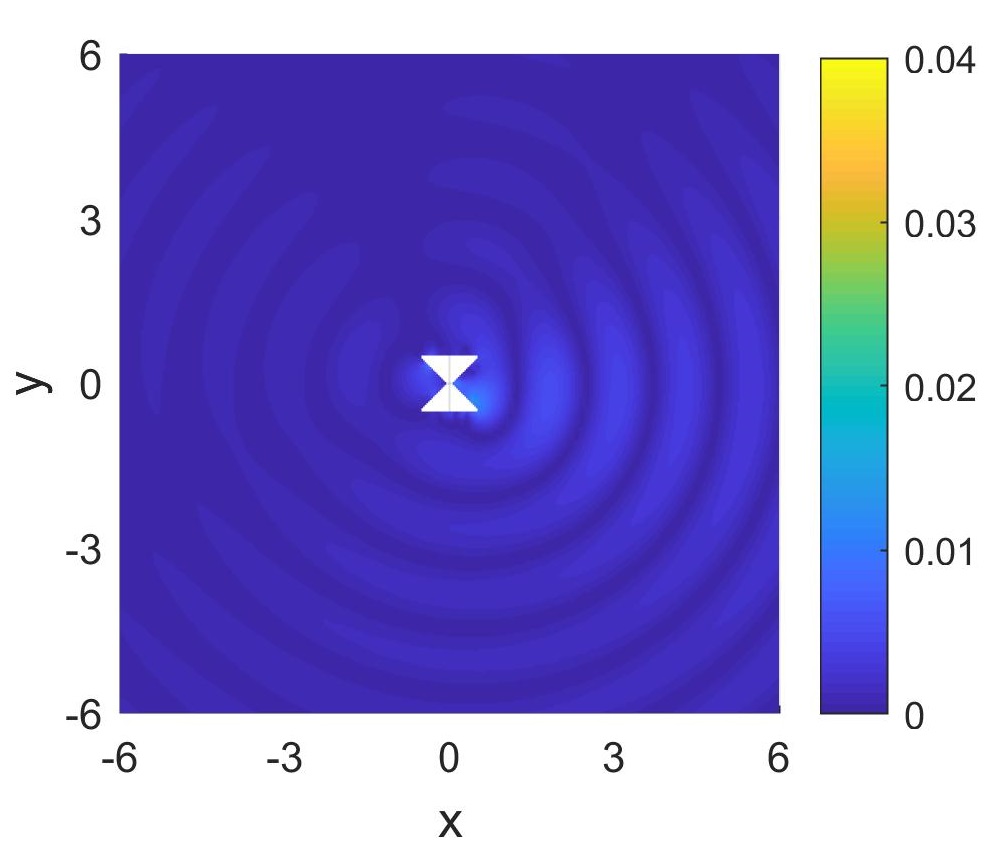}
%\caption{fig2}
}%
\caption{The first row show the results of $G(x,y)$ with $x=(-0.1163,-1.0780)$ and the second row show the it with $x=(0.3941,-0.1163)$ , (a) and (f) are the real part of the exact $G(x,y)$. (b) and (g) are real parts of numerical $G(x,y)$ respectively. (c),(d),(h) and (i) are corresponding imaginary part. (e) and (j) are the absolute errors between exact and numerical $G(x,y)$.}
\label{fig-ex2-1}
\end{figure}

In this experiment, let us consider a more practical scenario, a receiving antenna electromagnetic simulation problem. Assume we have a bow-tie antenna of 1 in length and 1 in width, which is a type of broad-bandwidth antenna. Its structure is 2 dimensional, implemented on a printed circuit board. Therefore, for simplicity, we consider simulating the field within the 2-D space. The shape of the bow-tie antenna can be seen in Figure \ref{fig-ex2-1}. We consider the Green's function of Helmholtz equation out of the bow-tie domains, and in this experiment, we consider the wavenumber $k=\pi-1/2$.

We also use the GreenNet formulation to compute the Green's function in the domain outside the bow-tie antenna. Because the domain is unbounded, we can only used the BI-GreenNet method. The neural network of this experiement has the similar architecture as the previous and the training process is also similar. We choose the ResNet architecture with 8 blocks and 100 neurons per layaer. We select 500 points equidistant on each edge of the boundary and randomly sample 100 points $x$ out of the domain $\Omega$ for calculating potential. After every 50 epochs of training, we will resample 100 points $x$. After training, we randomly sample the 100 $x$ and select $y$ at an interval of 0.1 in the domain $[-6,6]^2\backslash\Omega$. By computing the $G(x_i,y_j)$ on these point pairs $(x_i,y_j)$ we can find the relative $L^2$ error of the Green's function is 4.05e-2, which also achieve the high accuracy in exterior problem. When the wave number increases and the domain of the PDE problem becomes more complex, a high-precision Green's function approximation is still obtained. We randomly choose two points $x=(-0.1163,-1.0780)$ and $x=(0.3941,-0.1163)$, and show the results of $G(x,y)$ in Fig. \ref{fig-ex2-1}. We can see that compared with the exact solution, both the real part and the imaginary part have almost the same performance.

\subsubsection{The Helmholtz Equation out of a U-shape Domain}
\begin{figure}[t]
\centering
\subfigure[exact,real]{
\centering
\includegraphics[width=0.16\textwidth]{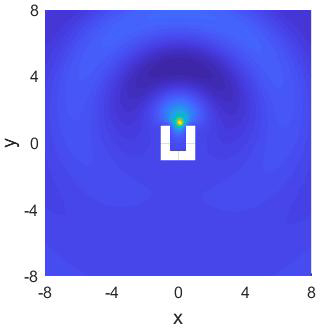}
%\caption{fig1}
}%
\subfigure[num,real]{
\centering
\includegraphics[width=0.21\textwidth]{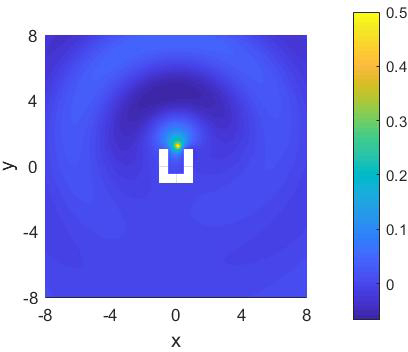}
%\caption{fig2}
}%
\subfigure[exact,imag]{
\centering
\includegraphics[width=0.16\textwidth]{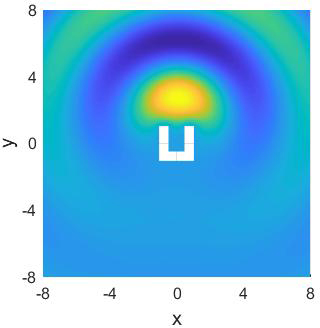}
%\caption{fig2}
}%
\subfigure[num,imag]{
\centering
\includegraphics[width=0.21\textwidth]{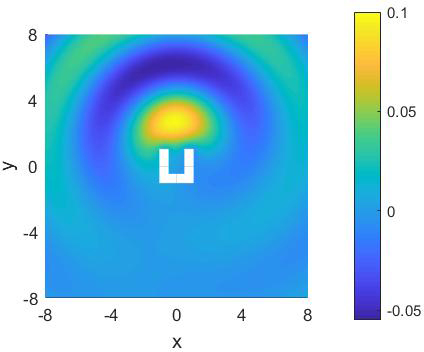}
%\caption{fig2}
}%
\subfigure[error]{
\centering
\includegraphics[width=0.21\textwidth]{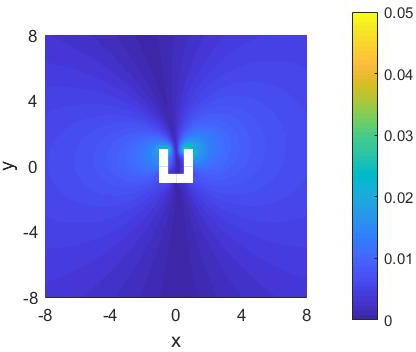}
%\caption{fig2}
}%

\subfigure[exact,real]{
\centering
\includegraphics[width=0.16\textwidth]{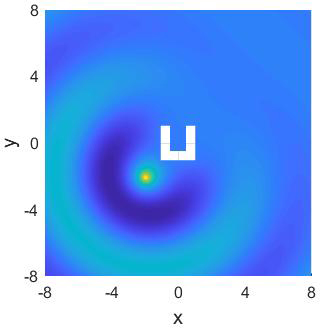}
%\caption{fig1}
}%
\subfigure[num,real]{
\centering
\includegraphics[width=0.21\textwidth]{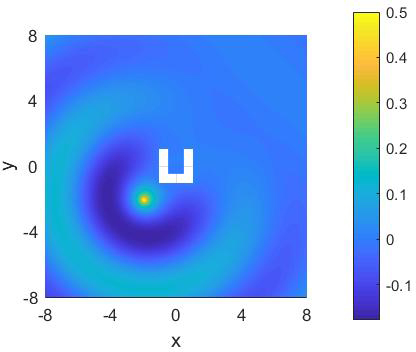}
%\caption{fig2}
}%
\subfigure[exact,imag]{
\centering
\includegraphics[width=0.16\textwidth]{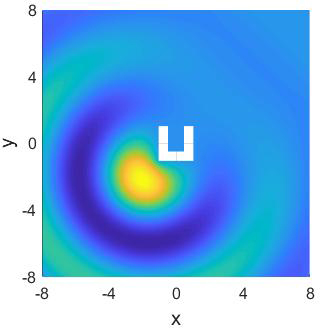}
%\caption{fig2}
}%
\subfigure[num,imag]{
\centering
\includegraphics[width=0.21\textwidth]{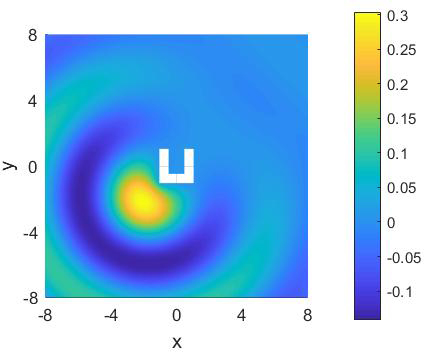}
%\caption{fig2}
}%
\subfigure[error]{
\centering
\includegraphics[width=0.21\textwidth]{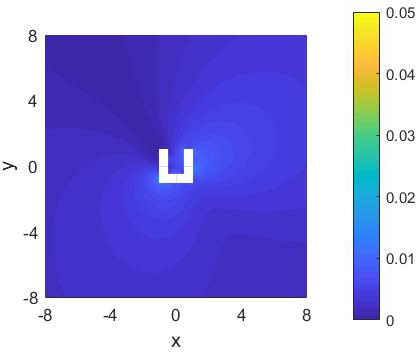}
%\caption{fig2}
}%

\caption{This figure show the results of Green's function out of the U-shaped domain. The first row are the results of $G(x,y)$ with fixed $x=(0.2,1.2)$ and the second row are the results of $G(x,y)$ with fixed $x=(-1.9689 -2.0929)$. The exact Green's function are also shown for comparison.}
\label{fig-ex2-2}
\end{figure}

The U-shaped domain is also a common consideration area. In this experiment, we assume the wavenumber $k=1$ and the setting of the domain is shown in Fig. \ref{fig-ex2-2}. We also used the BI-GreenNet with a ResNet architecture network that has 8 blocks and 100 neurons per layer. 
We select 400 points equidistant on each edge of the boundary and randomly sample 150 points $x$ out of the domain $\Omega$ for calculating potential. After every 100 epochs of training, we will resample 150 points $x$.

After training, we randomly sample the 200 $x$s and select $y$ at an interval of 0.4 in the domain $[-8,8]^2\backslash\Omega$. By computing the $G(x_i,y_j)$ on these point pairs, the relative $L^2$ error of the Green's function is 2.06e-2, which also achieve the high accuracy. We choose two points $x=(0.2,1.2)$ and $x=(-1.9689 -2.0929)$ and show the $G(x,y)$ of the two points in Fig. \ref{fig-ex2-2}. We can also see that both the real part and the imaginary part are very close to the exact solution.

\subsection{The Green's function of the interface problem}
\subsubsection{The Poisson equation in a square with flower-shaped interface}
Let $\Omega$ be the square $\{(x,y): \vert x\vert\leq1, \vert y\vert\leq 1\}$, the interface $\Gamma$ is parameterized by $(a\cos t-b\cos nt\cos t, a\sin t-b\cos nt\sin t)$ with $t\in[0,2\pi]$, which is a flower-shaped interface widely used in \cite{zhao2010high, marques2011correction}. We choose two set of parameters $(a,b,n)=(0.5,0.15,5)$ and $(a,b,n)=(0.5,0.1,8)$. For the first parameter, we choose $\mu_1=1, \mu_2=2$ while for the second parameter, we take $\mu_1=2, \mu_2=1$, as is shown in Fig. \ref{pint}.

\begin{figure}[htbp]
\centering
\subfigure[$(a,b,n)=(0.5,0.15,5)$]{
\centering
\includegraphics[width=0.4\textwidth]{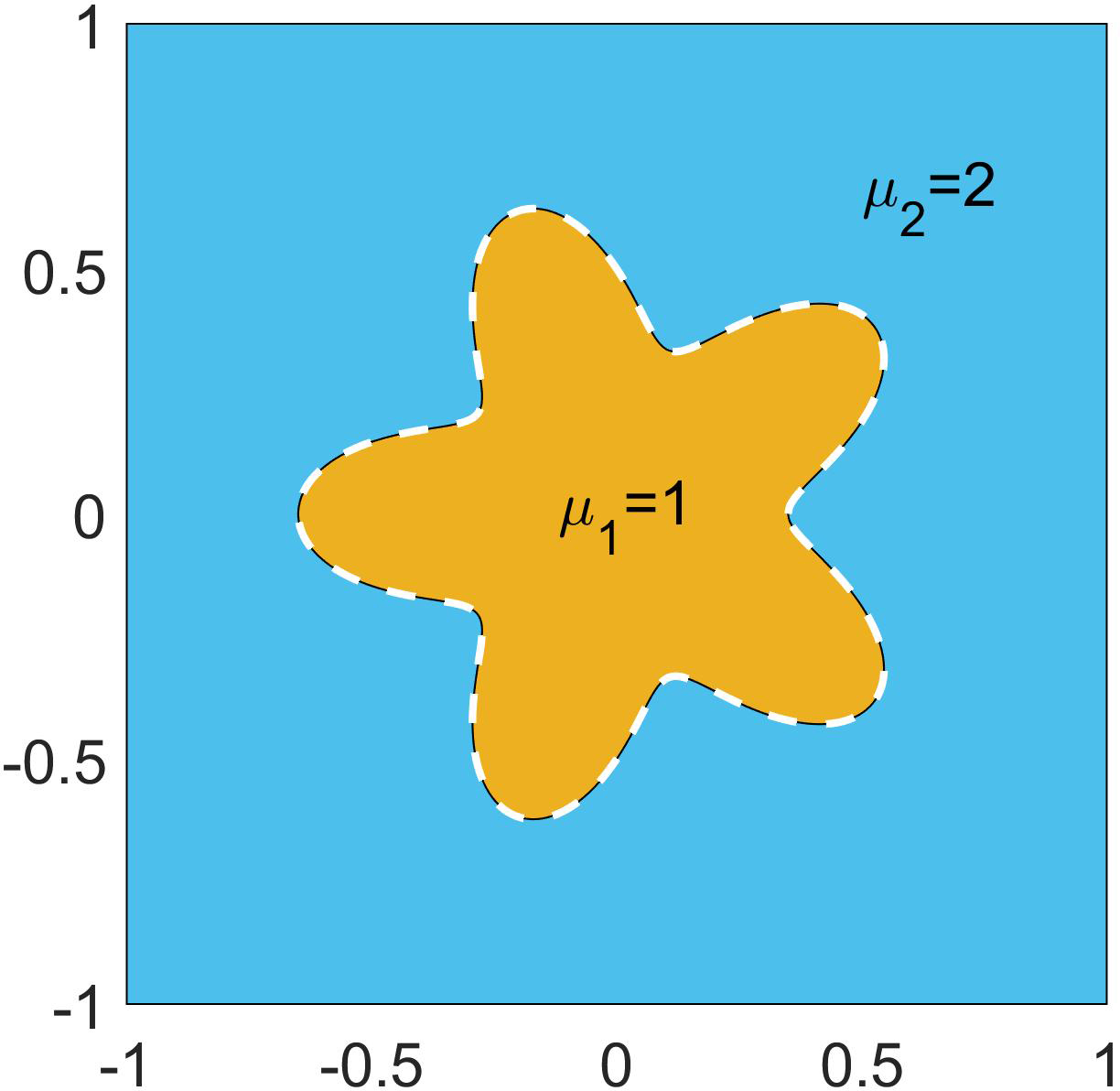}
%\caption{fig1}
}%
\hspace{0.2in}
\subfigure[$(a,b,n)=(0.5,0.1,8)$]{
\centering
\includegraphics[width=0.4\textwidth]{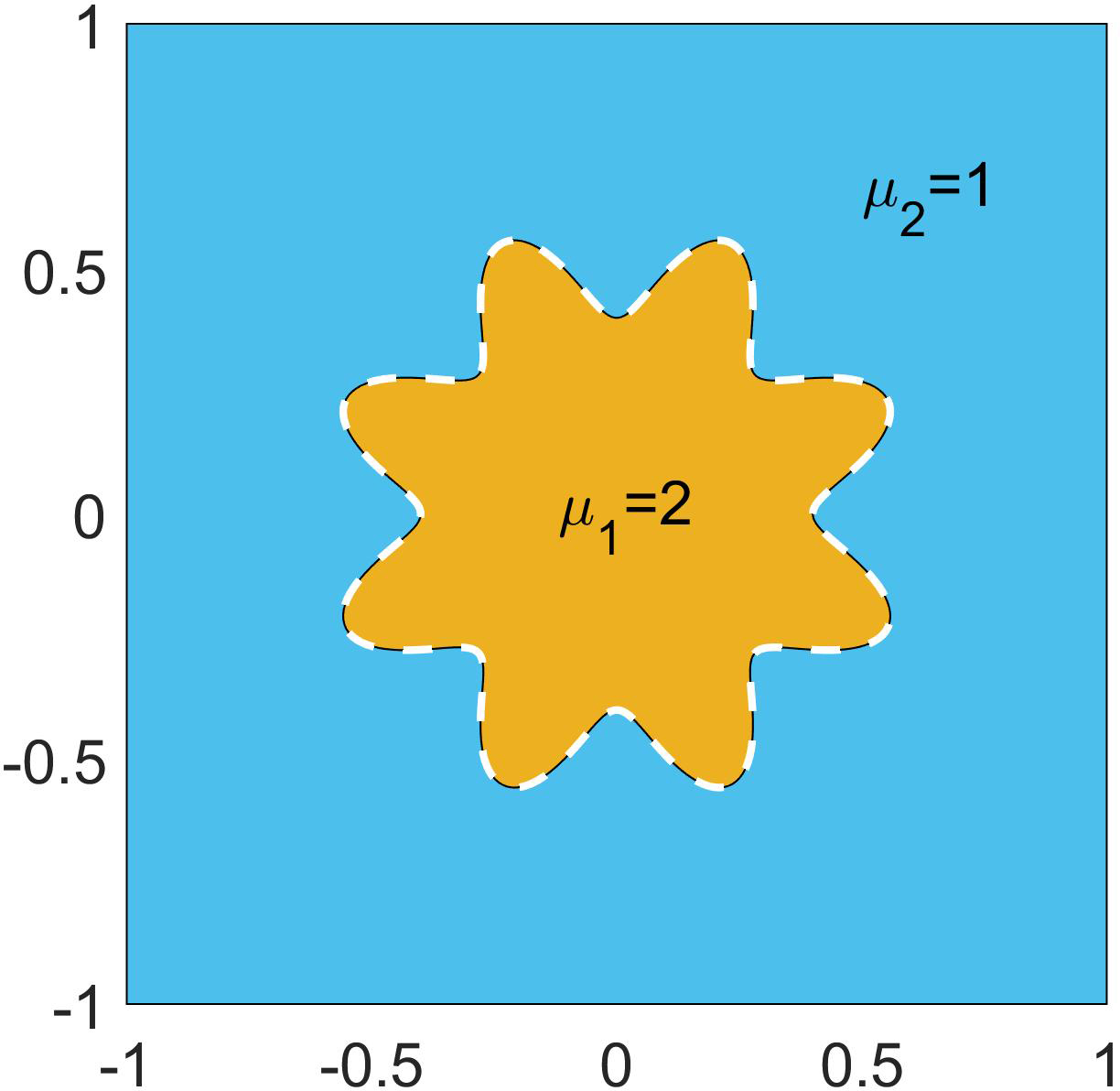}}
\caption{Illustration of the Poission interface problem}
\label{pint}
\end{figure}

We impose Dirichlet boundary condition on $\partial\Omega$ such that the Green's function to this problem satisfies (\ref{G_inter}) along with $G(x,y)=0, \ \forall x\in\Omega, y\in\partial\Omega$. We sample 800 points on both $\Gamma$ and $\partial\Omega$ for boundary integral. In the training process, we choose a ResNet with 8 blocks with 100 neurons per layer. For every 500 epochs, 100 new $x$ is randomly generated. After after $10^5$ epochs of training, we fix 100 newly generated $x$ and compute the average relative $L_2$ error of the Green's function $G(x,y)$. Further more, for two fixed $x$ in $\Omega_1$ and $\Omega_2$ respectively, we compare the exact Green's function (obtained by traditional boundary integral method) and the numerical solution obtained by neural network.

\begin{figure}[htbp]
\centering
\subfigure[Exact solution]{
\centering
\includegraphics[width=0.3\textwidth]{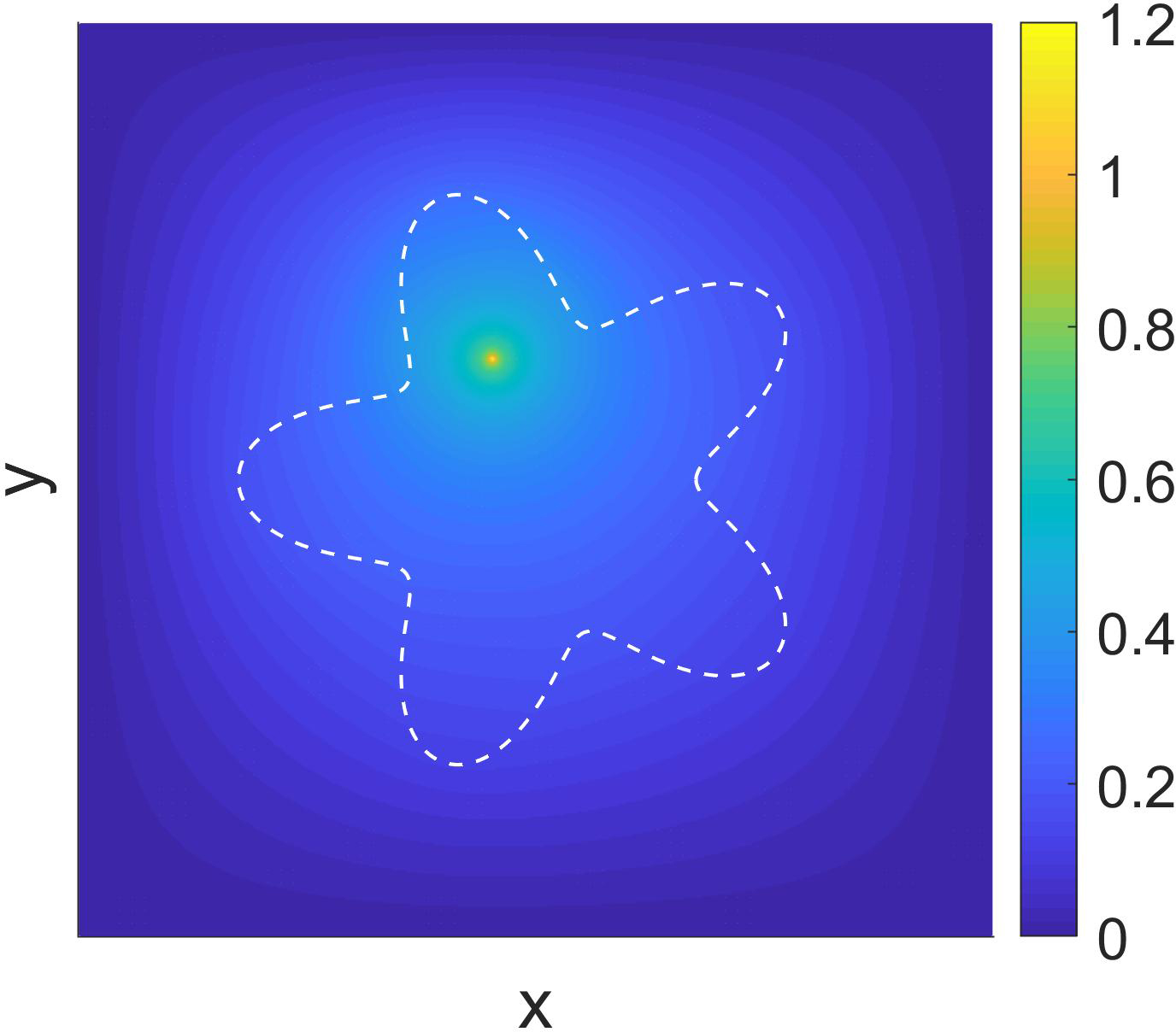}
%\caption{fig1}
}%
\subfigure[Numerical solution]{
\centering
\includegraphics[width=0.3\textwidth]{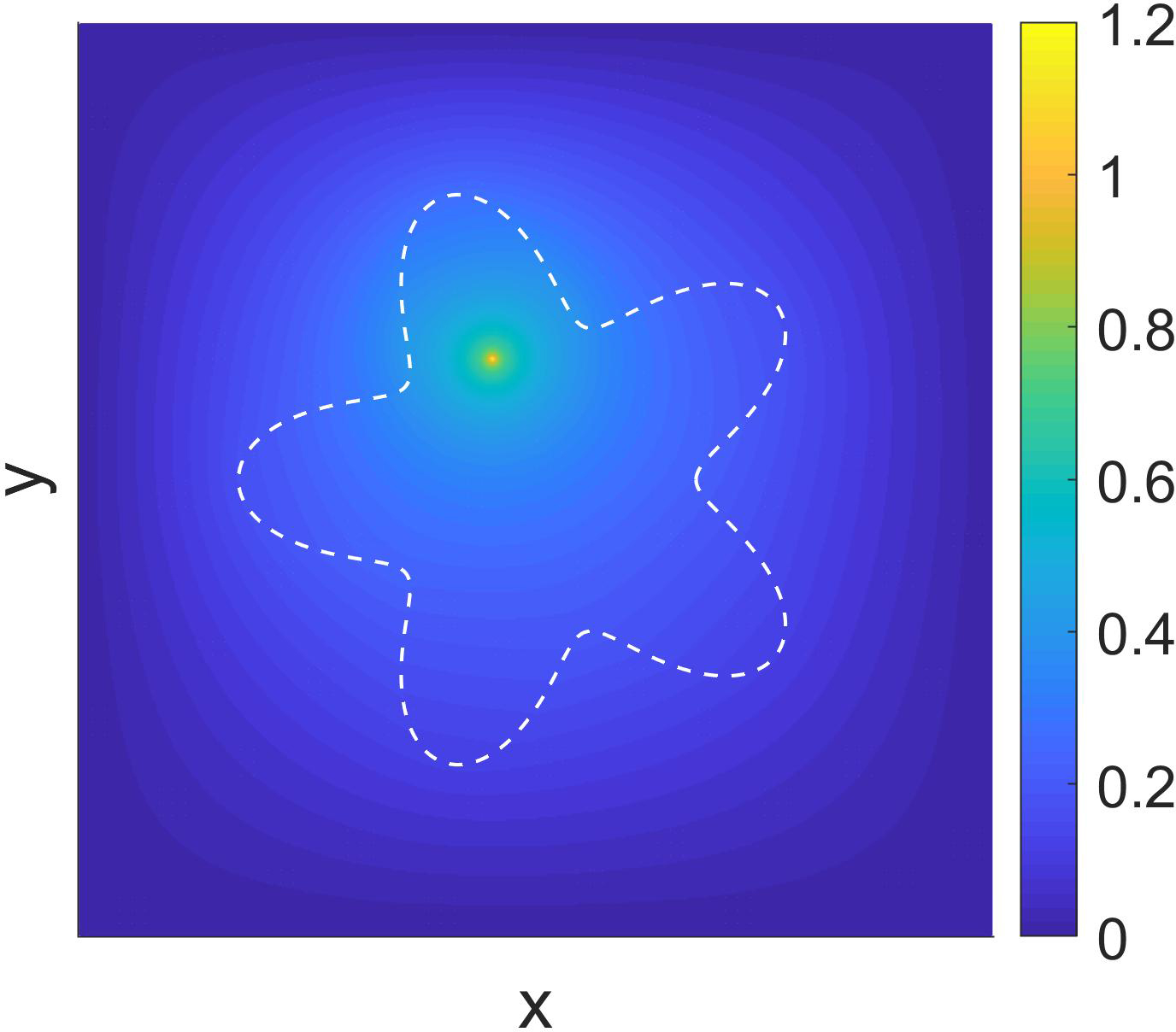}
%\caption{fig2}
}%
\subfigure[Absolute error]{
\centering
\includegraphics[width=0.32\textwidth]{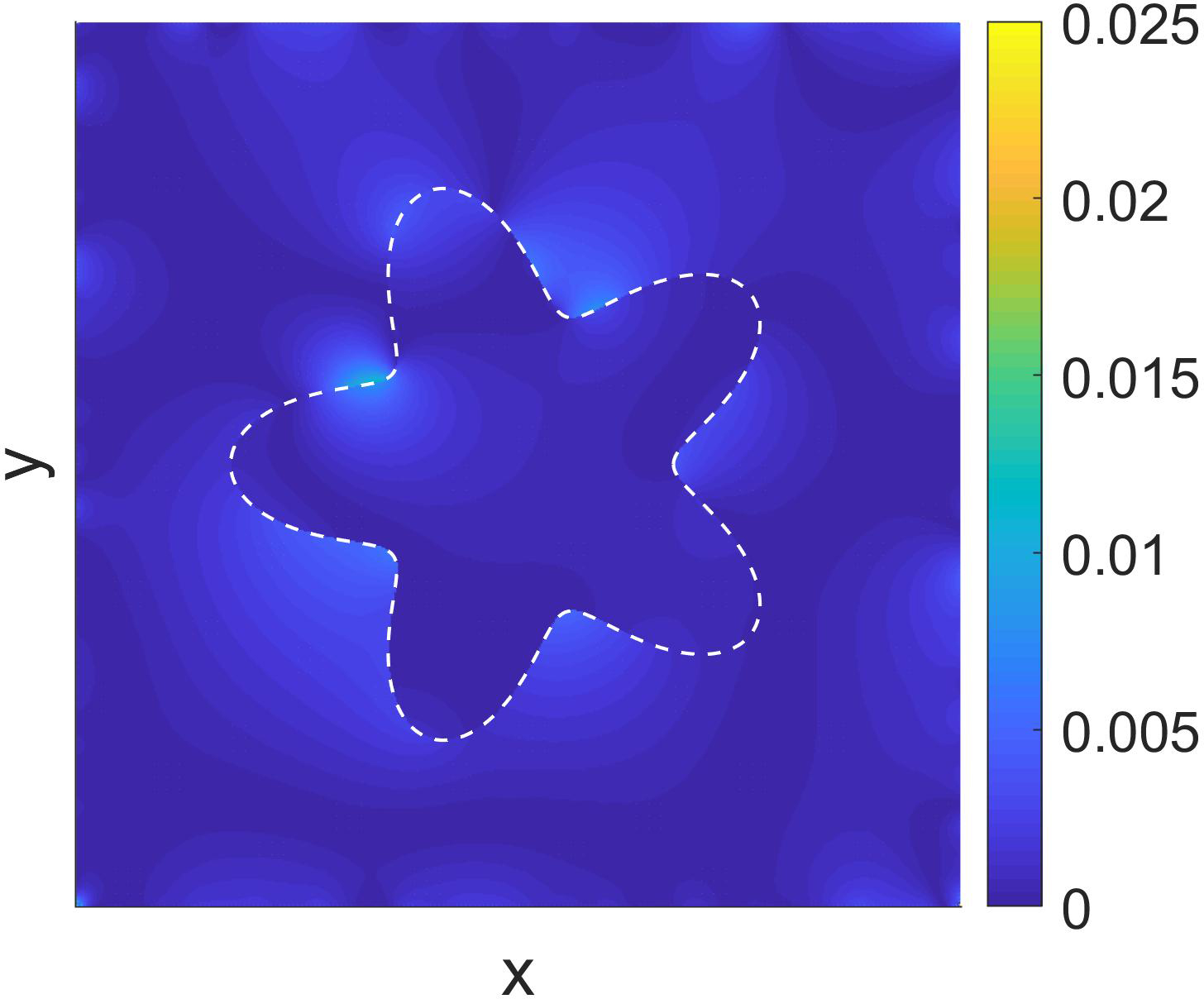}
}
\subfigure[Exact solution]{
\centering
\includegraphics[width=0.3\textwidth]{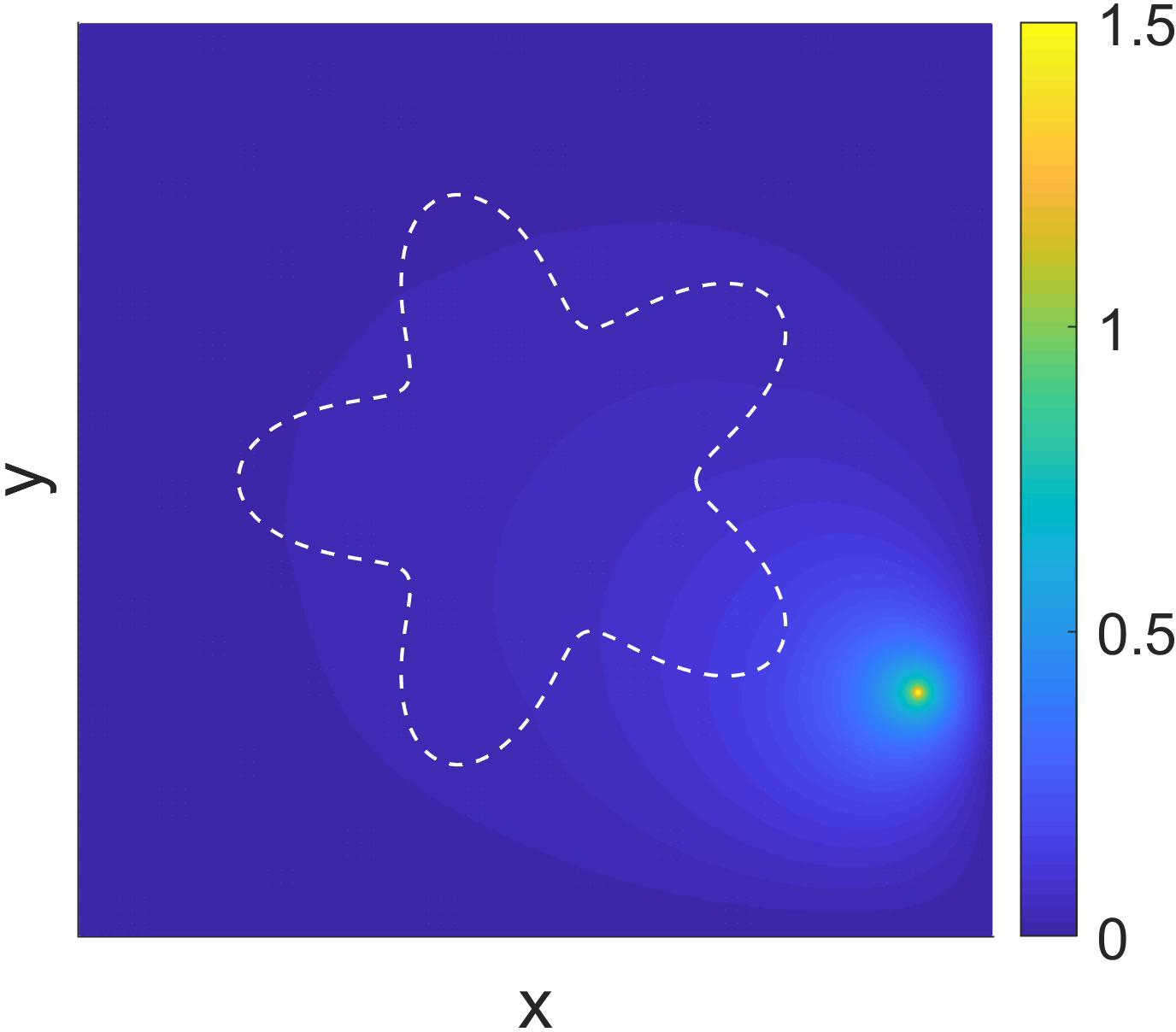}
%\caption{fig1}
}%
\subfigure[Numerical solution]{
\centering
\includegraphics[width=0.3\textwidth]{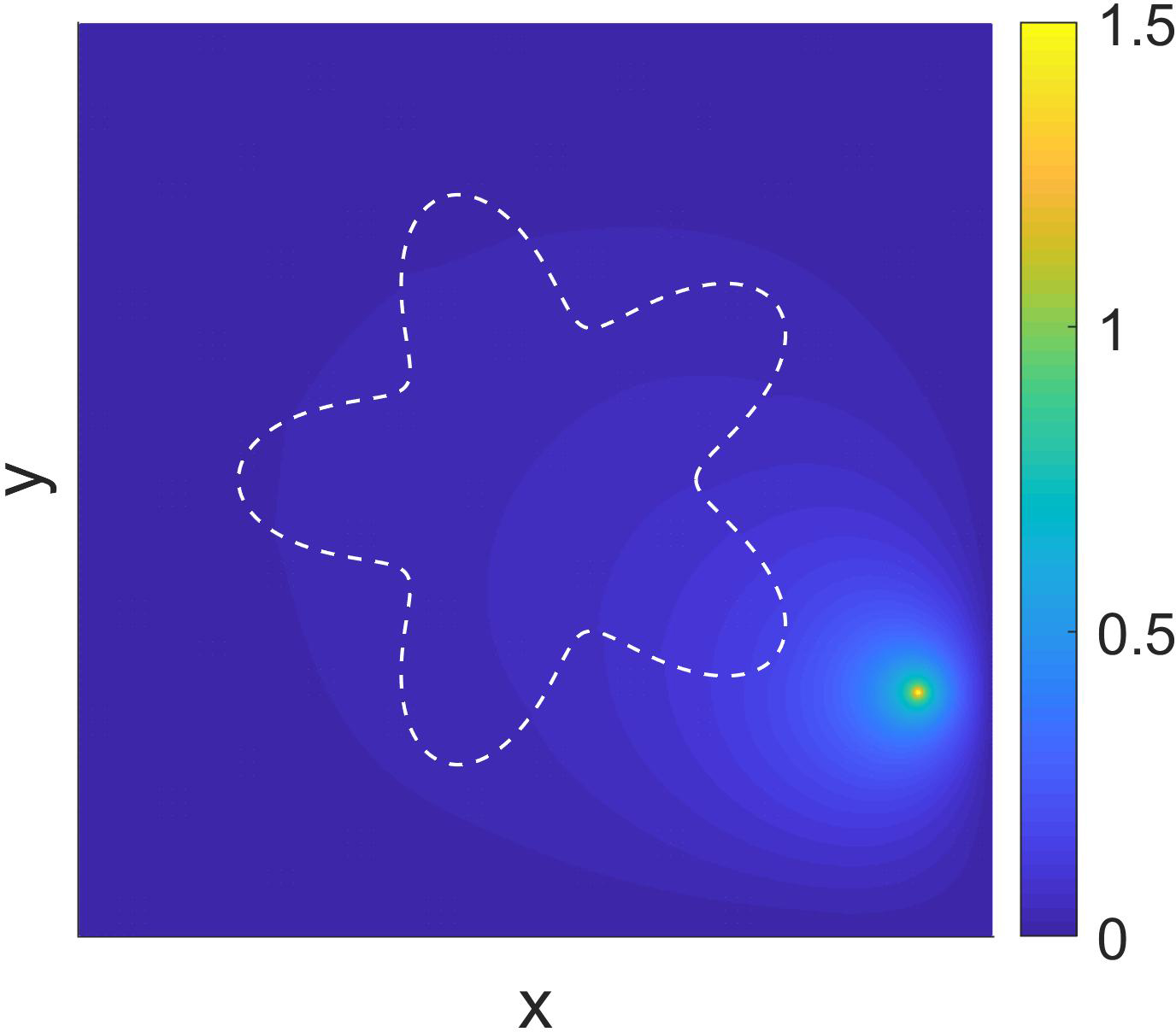}
%\caption{fig2}
}%
\subfigure[Absolute error]{
\centering
\includegraphics[width=0.32\textwidth]{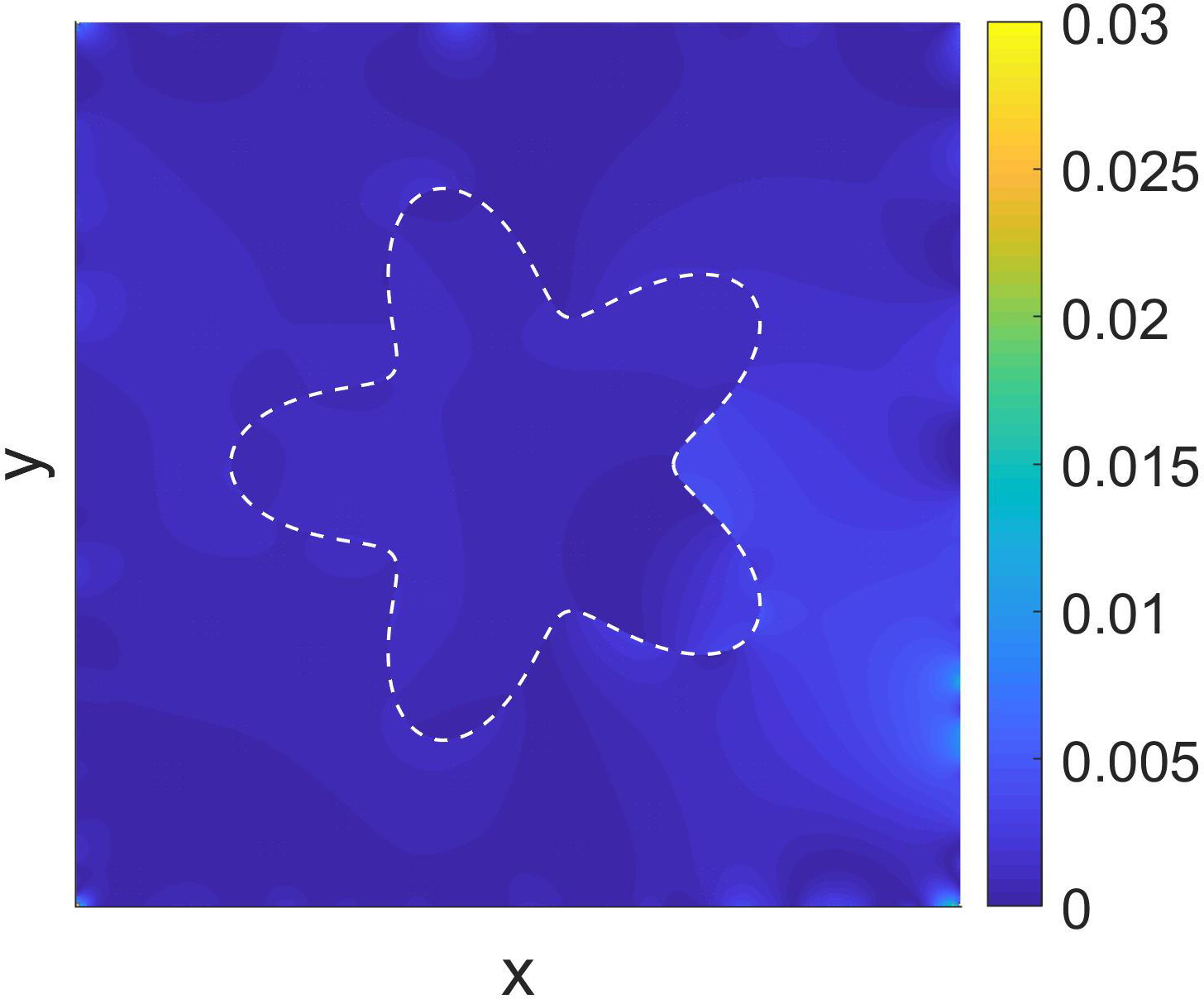}
}
\caption{Numerical results for the Poisson interface problem with $(a,b,n)=(0.5,0.15,5)$ and $\mu_1=1, \mu_2=2$. Figures (a)-(c): exact solution, numerical solution and absolute error of the Green's function $G(x,y)$ for $x=(-0.0960,0.2626)\in\Omega_1$. Figures (d)-(f):  exact solution, numerical solution and absolute error of the Green's function $G(x,y)$ for $x=(0.8342,-0.4661)\in\Omega_2$.}
\label{fig-ex4-1}
\end{figure}

\begin{figure}[htbp]
\centering
\subfigure[exact solution]{
\centering
\includegraphics[width=0.3\textwidth]{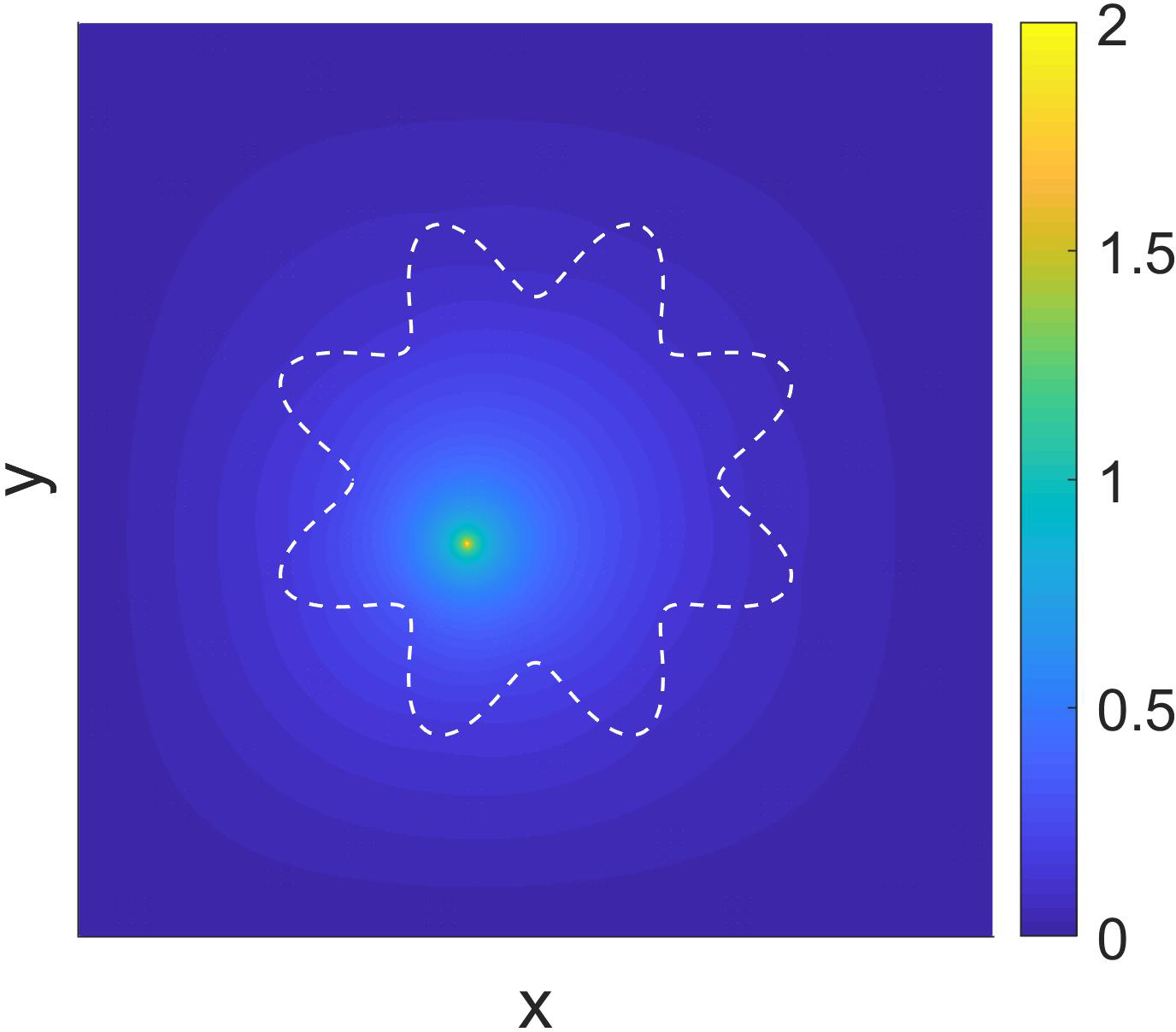}
%\caption{fig1}
}%
\subfigure[numerical solution]{
\centering
\includegraphics[width=0.3\textwidth]{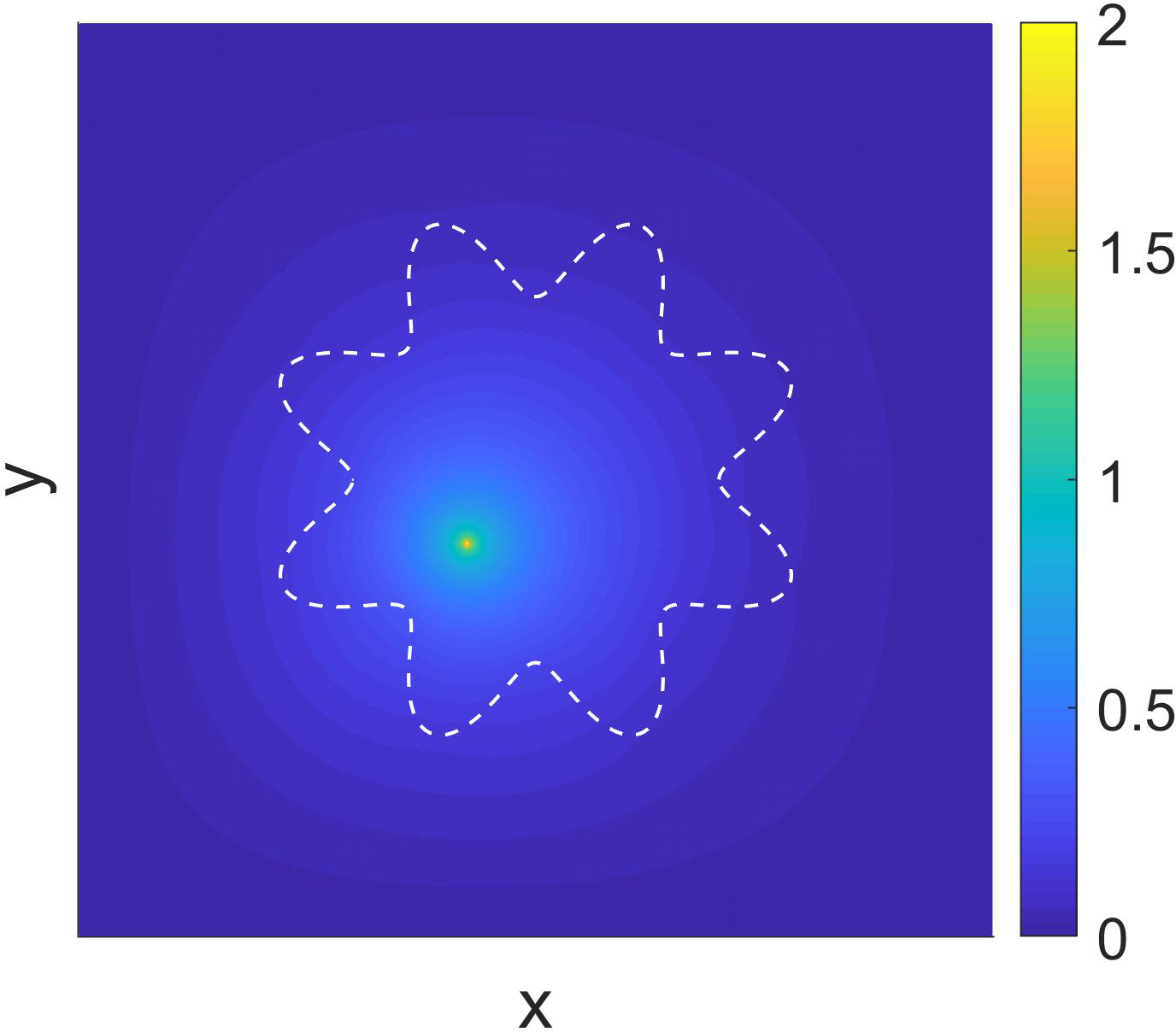}
%\caption{fig2}
}%
\subfigure[absolute error]{
\centering
\includegraphics[width=0.32\textwidth]{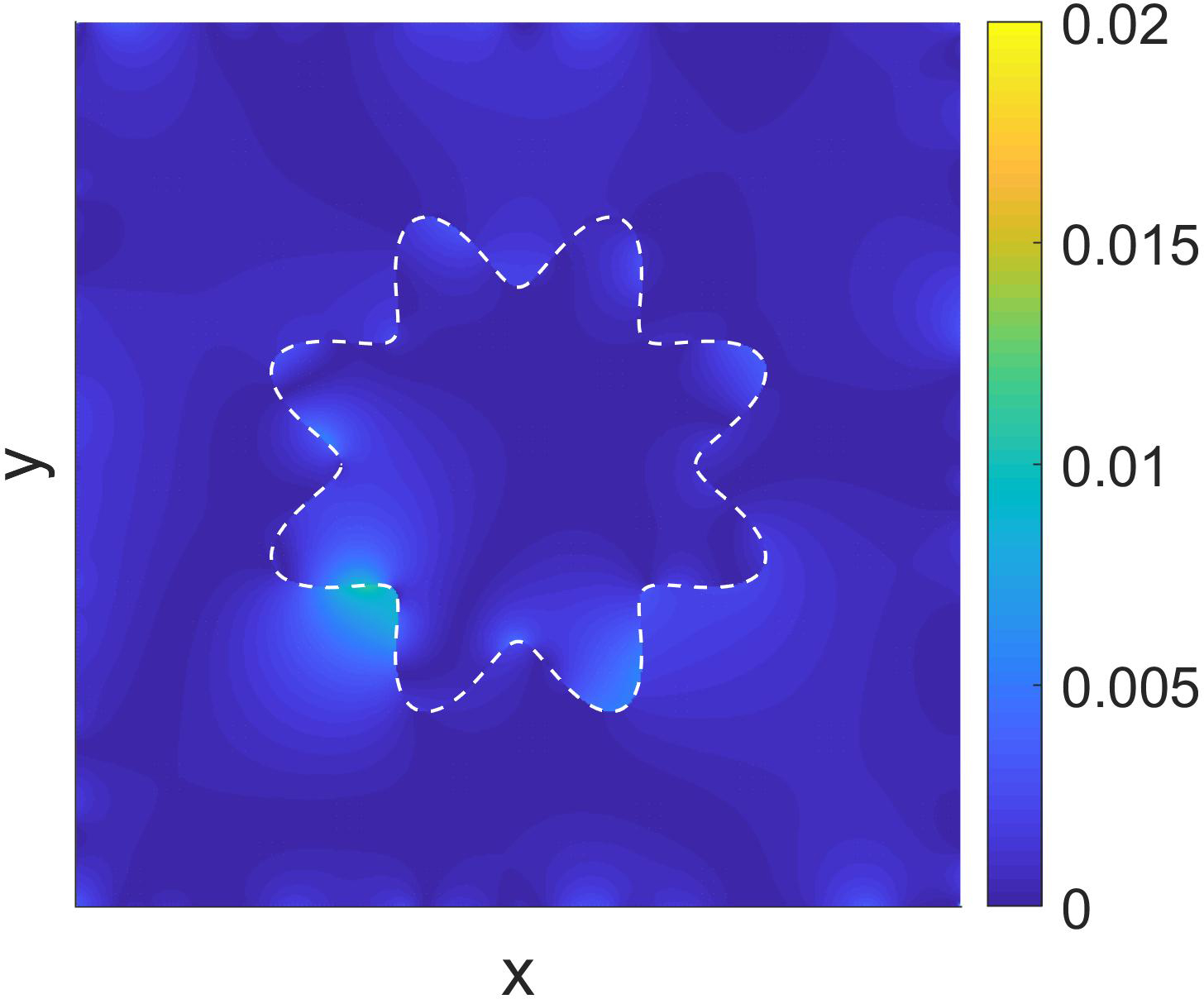}
}
\subfigure[exact solution]{
\centering
\includegraphics[width=0.3\textwidth]{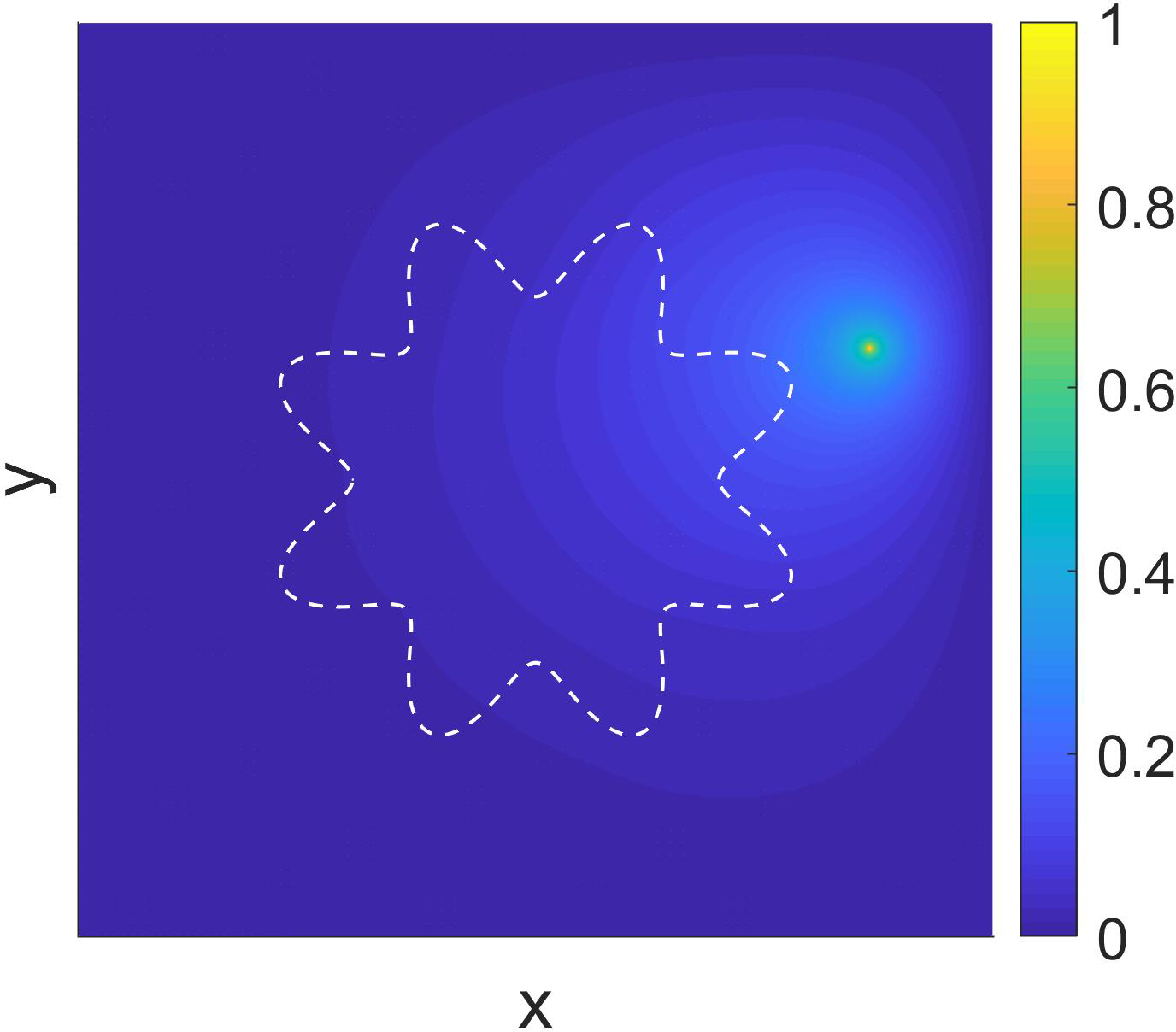}
%\caption{fig1}
}%
\subfigure[numerical solution]{
\centering
\includegraphics[width=0.3\textwidth]{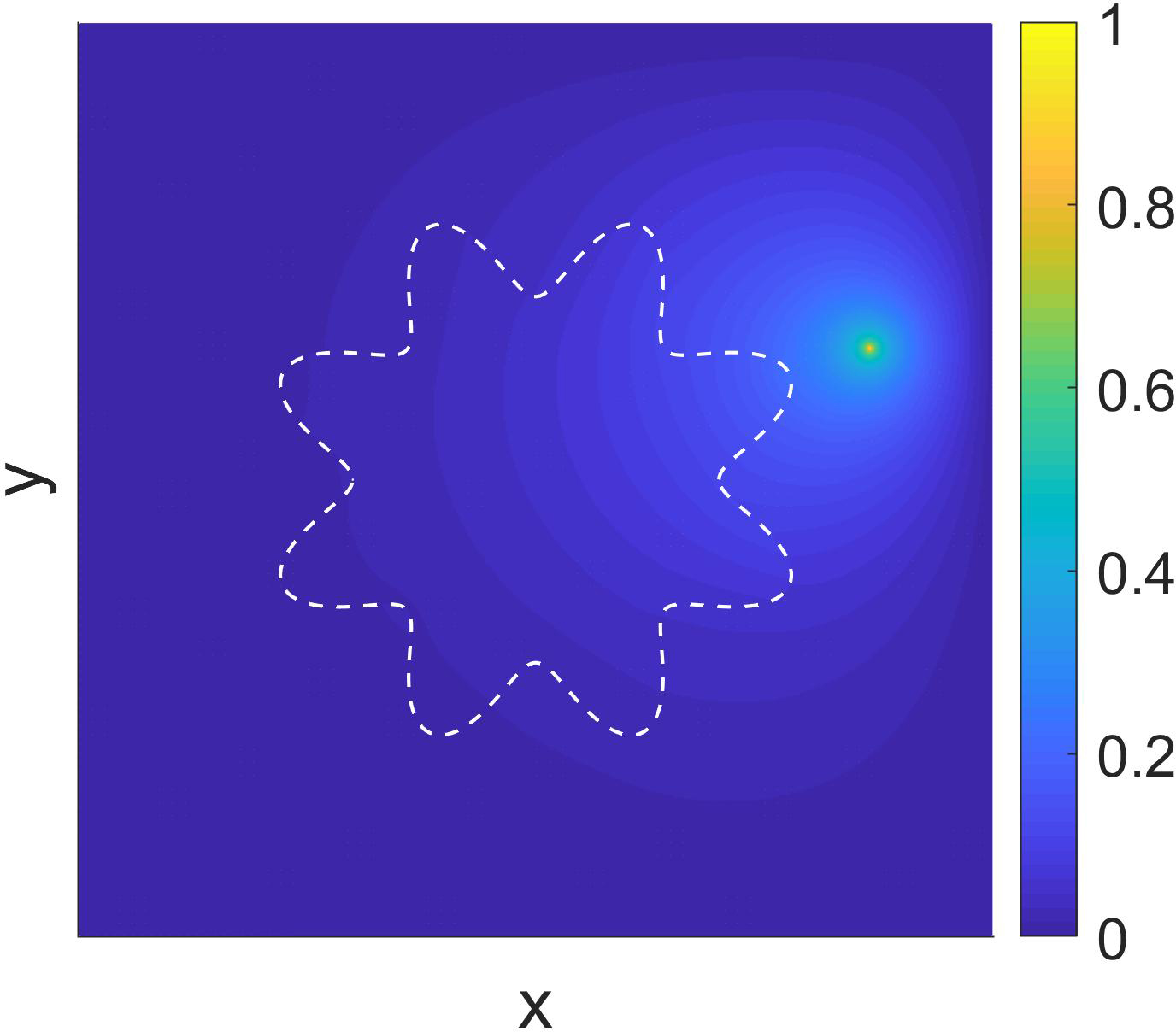}
%\caption{fig2}
}%
\subfigure[absolute error]{
\centering
\includegraphics[width=0.32\textwidth]{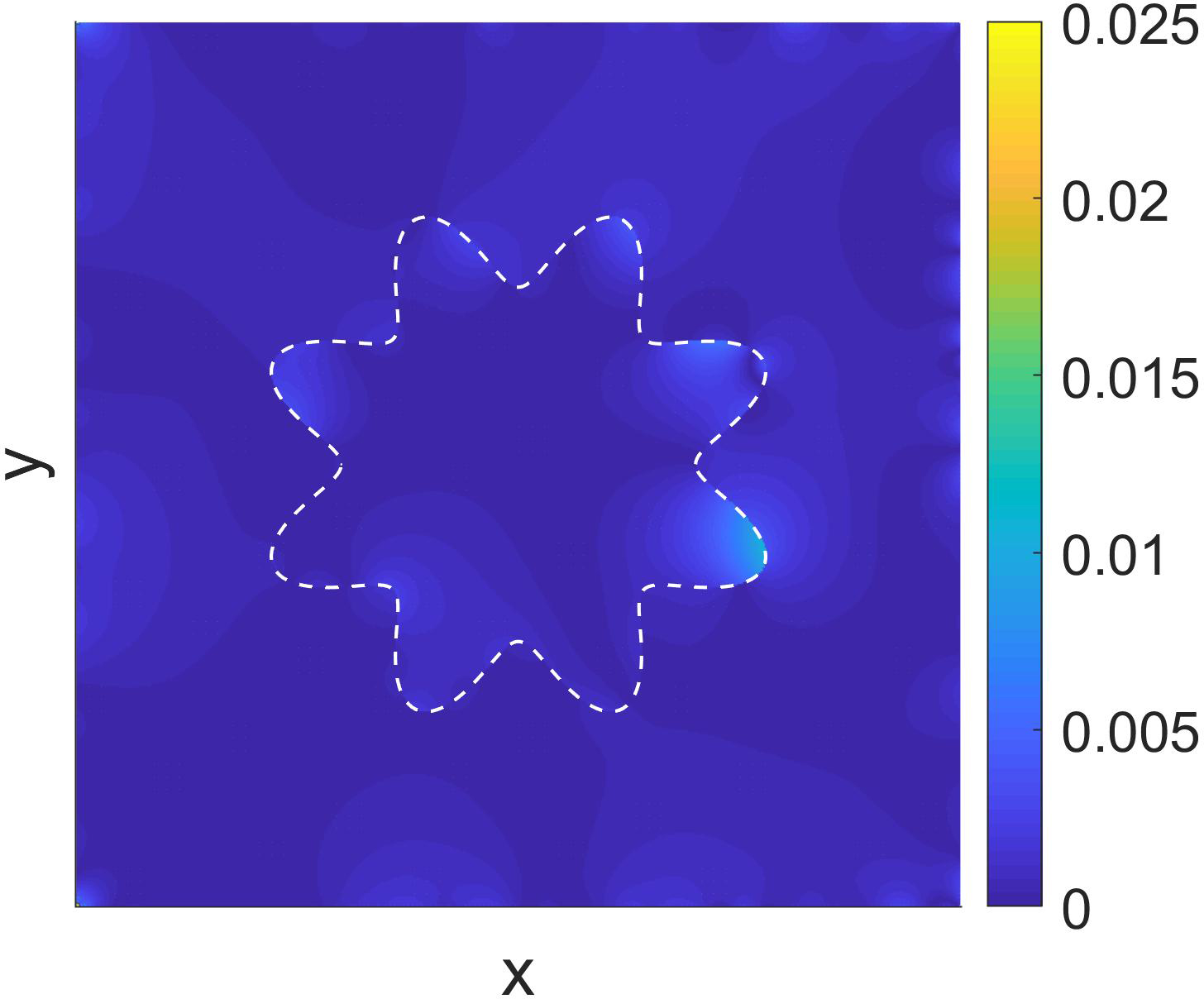}
}
\caption{Numerical results for the Poisson interface problem with $(a,b,n)=(0.5,0.1,8)$ and $\mu_1=2, \mu_2=1$. Figures (a)-(c): exact solution, numerical solution and absolute error of the Green's function $G(x,y)$ for $x=(-0.1512,-0.1410)\in\Omega_1$. Figures (d)-(f):  exact solution, numerical solution and absolute error of the Green's function $G(x,y)$ for $x=(0.7288,0.2861)\in\Omega_2$.}
\label{fig-ex4-2}
\end{figure}

The numerical results for the second set of parameters is given in Fig.\ref{fig-ex4-1}. The average relative $L_2$ error of the Green's function $G(x,y)$ is $1.85\times10^{-2}$. For $x=(-0.0960,0.2626)\in\Omega_1$, the relative $L^2$ error of the Green's function is $0.78\times10^{-2}$, while for $x=(0.8342,-0.4661)\in\Omega_2$, the relative $L^2$ error is $1.79\%$. The numerical results for the second set of parameters is given in Fig.\ref{fig-ex4-2}. The average relative $L_2$ error of the Green's function $G(x,y)$ is $3.34\times10^{-2}$. For $x=(-0.1512,-0.1410)\in\Omega_1$, the relative $L^2$ error of the Green's function is $0.81\times10^{-2}$, while for $x=(0.7288,0.2861)\in\Omega_2$, the relative $L^2$ error is $1.37\times10^{-2}$. The results both show that the proposed method can solve the Green's function for the interface problem accurately.

\subsubsection{The Helmholtz equation in $\mathbb{R}^2$ with square interface}
In this experiment, we consider the interface problem of the Helmholtz equation \eqref{interface} with the interface $\Gamma$ = $\{(x,y): \vert x\vert=1 \ \text{or} \ \vert y\vert=1\}$. On the interface 800 points are sampled for the boundary integral. Take $\mu_1=2, \mu_2=1, \varepsilon_1=1, \varepsilon_2=4, k=2$ in the Helmholtz equation. The Sommerfeld condition is 
required at infinity, i.e., $\lim_{\vert x\vert\to\infty}(\frac{\partial}{\partial r}-ik\sqrt{\varepsilon_2\mu_2})u(x)=o(\vert x\vert^{-1/2})$, implying that $\lim_{\vert y\vert\to\infty}(\frac{\partial}{\partial r}-ik\sqrt{\varepsilon_2\mu_2})G(x,y)=o(\vert x\vert^{-1/2})$, which is automatically satisfied if $H(x,y)$ is written in the boundary integral form. Therefore, only the two jump conditions need to be fit since the partial differential equation is already satisfied using the boundary integral representation.

\begin{figure}[htbp]
\centering
\subfigure[Exact, real]{
\centering
\includegraphics[width=0.19\textwidth]{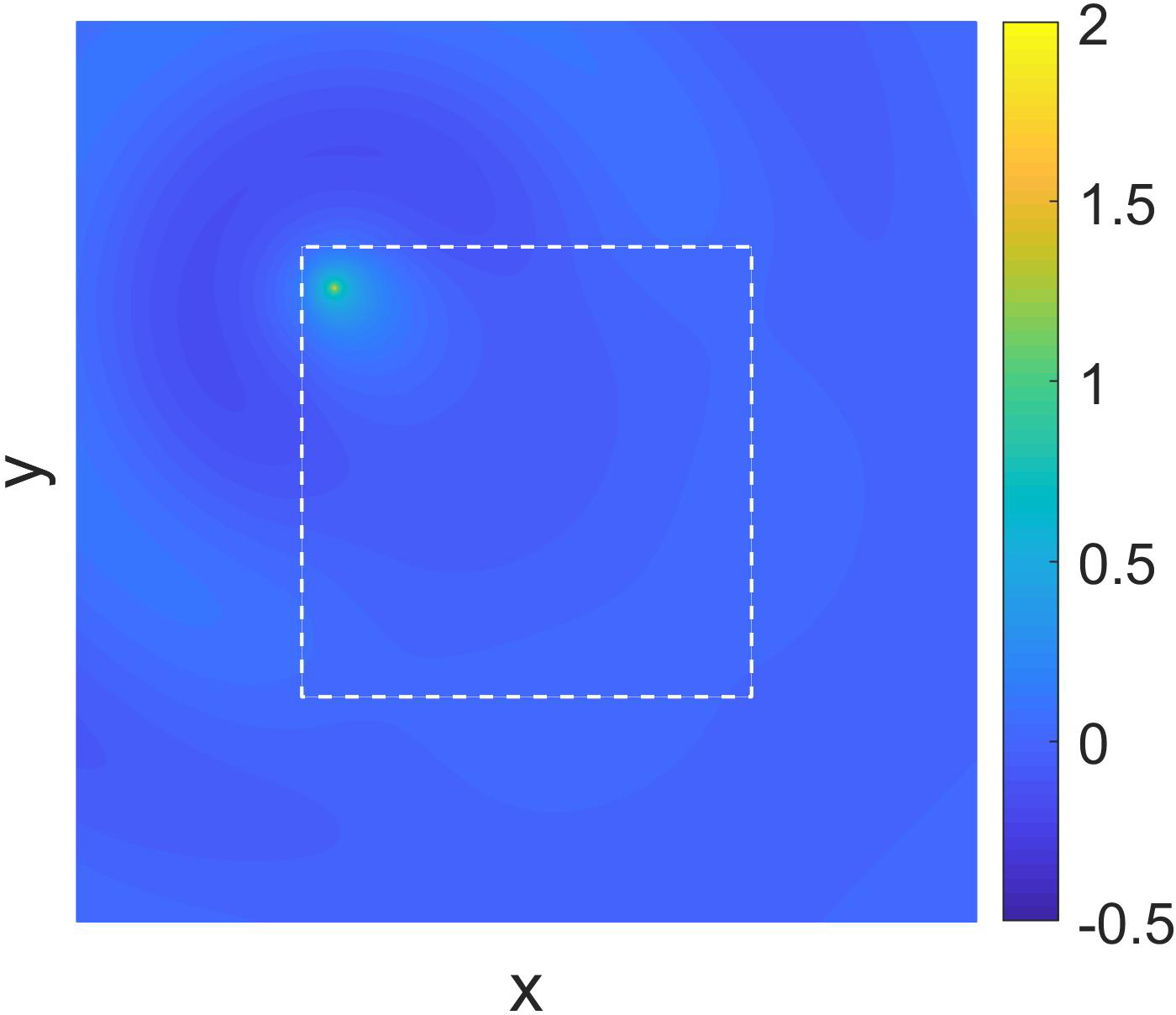}
%\caption{fig1}
}%
\subfigure[Num, real]{
\centering
\includegraphics[width=0.19\textwidth]{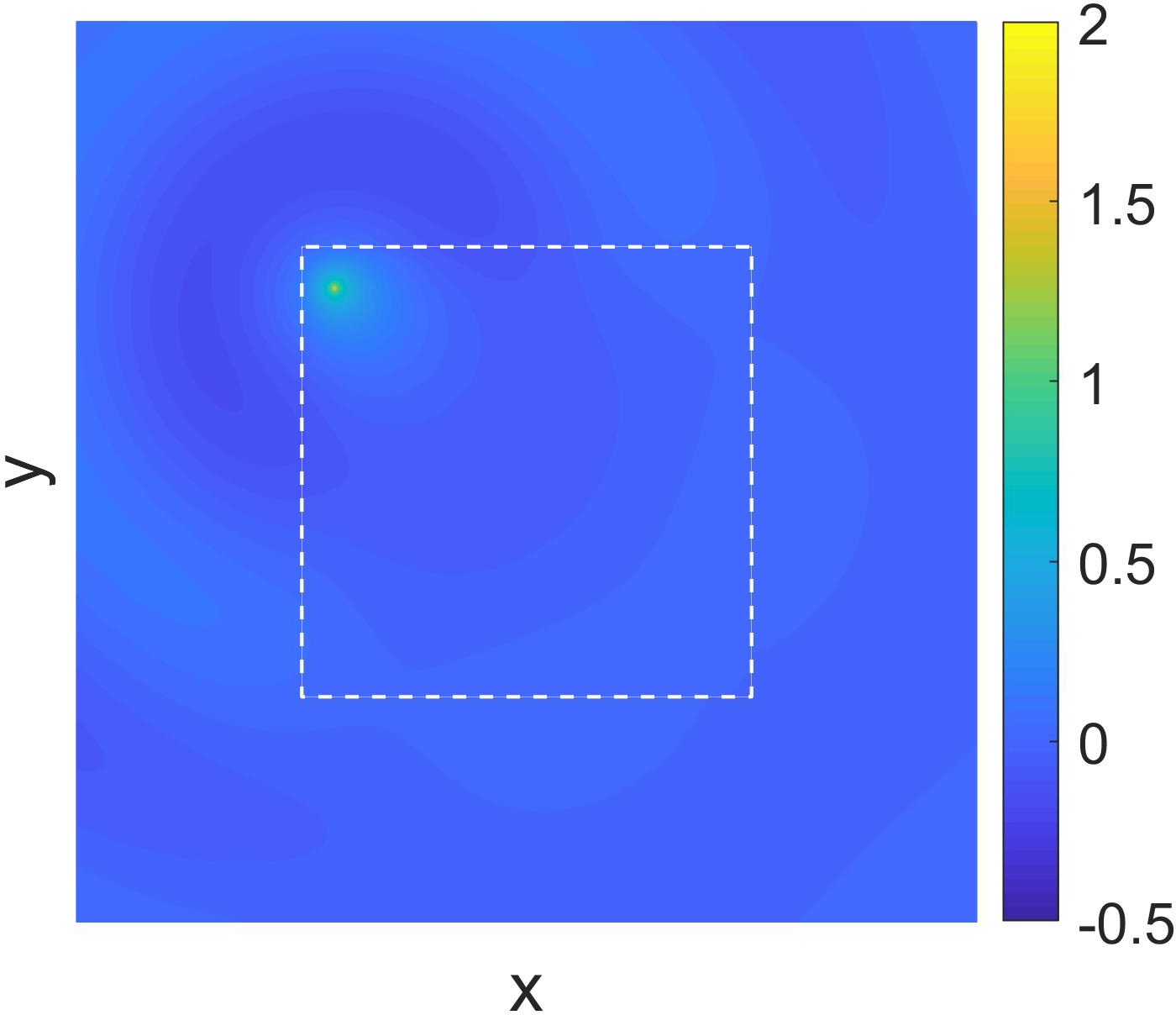}
%\caption{fig2}
}%
\subfigure[Exact, imag]{
\centering
\includegraphics[width=0.19\textwidth]{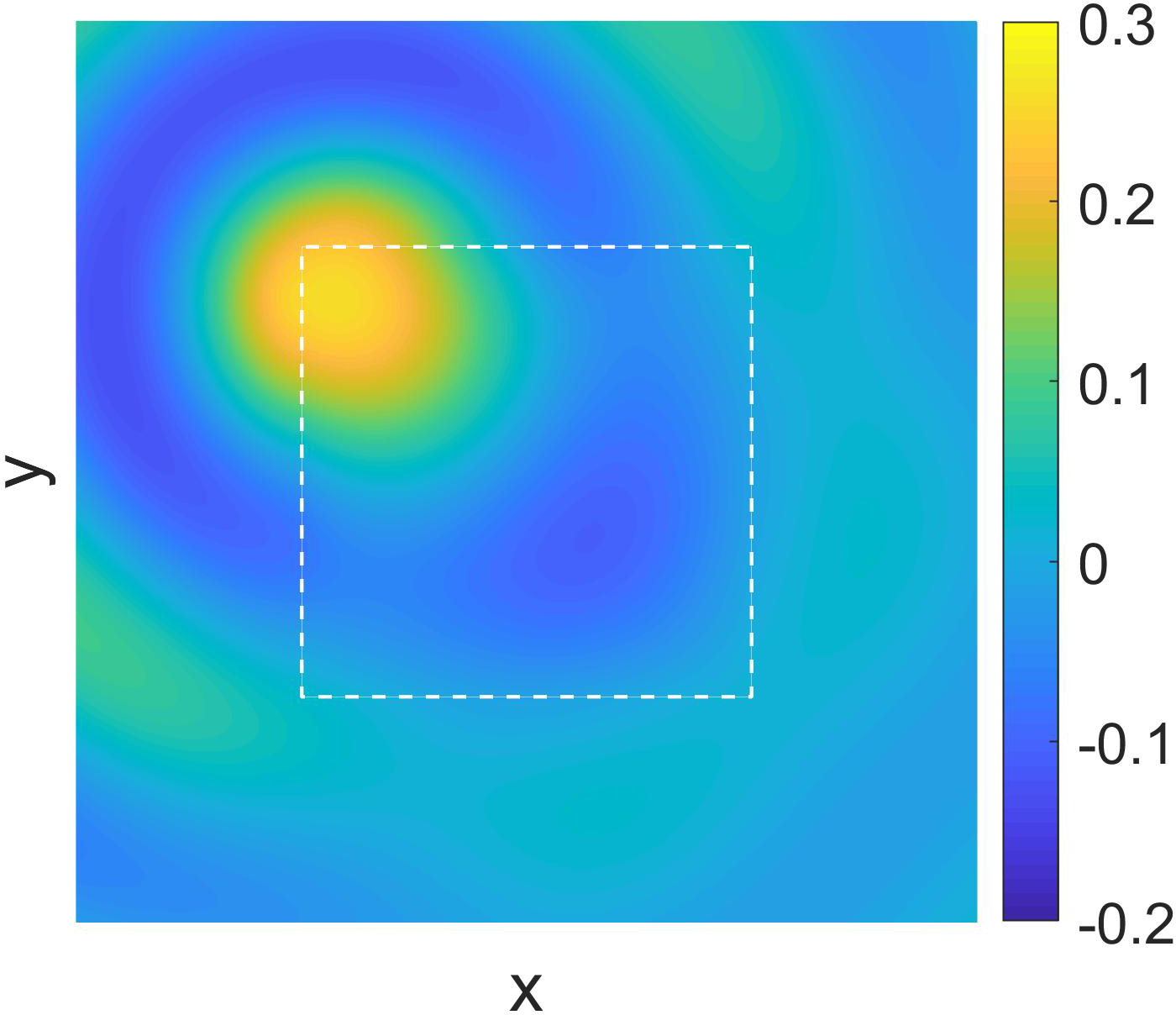}
%\caption{fig2}
}%
\subfigure[Num, imag]{
\centering
\includegraphics[width=0.19\textwidth]{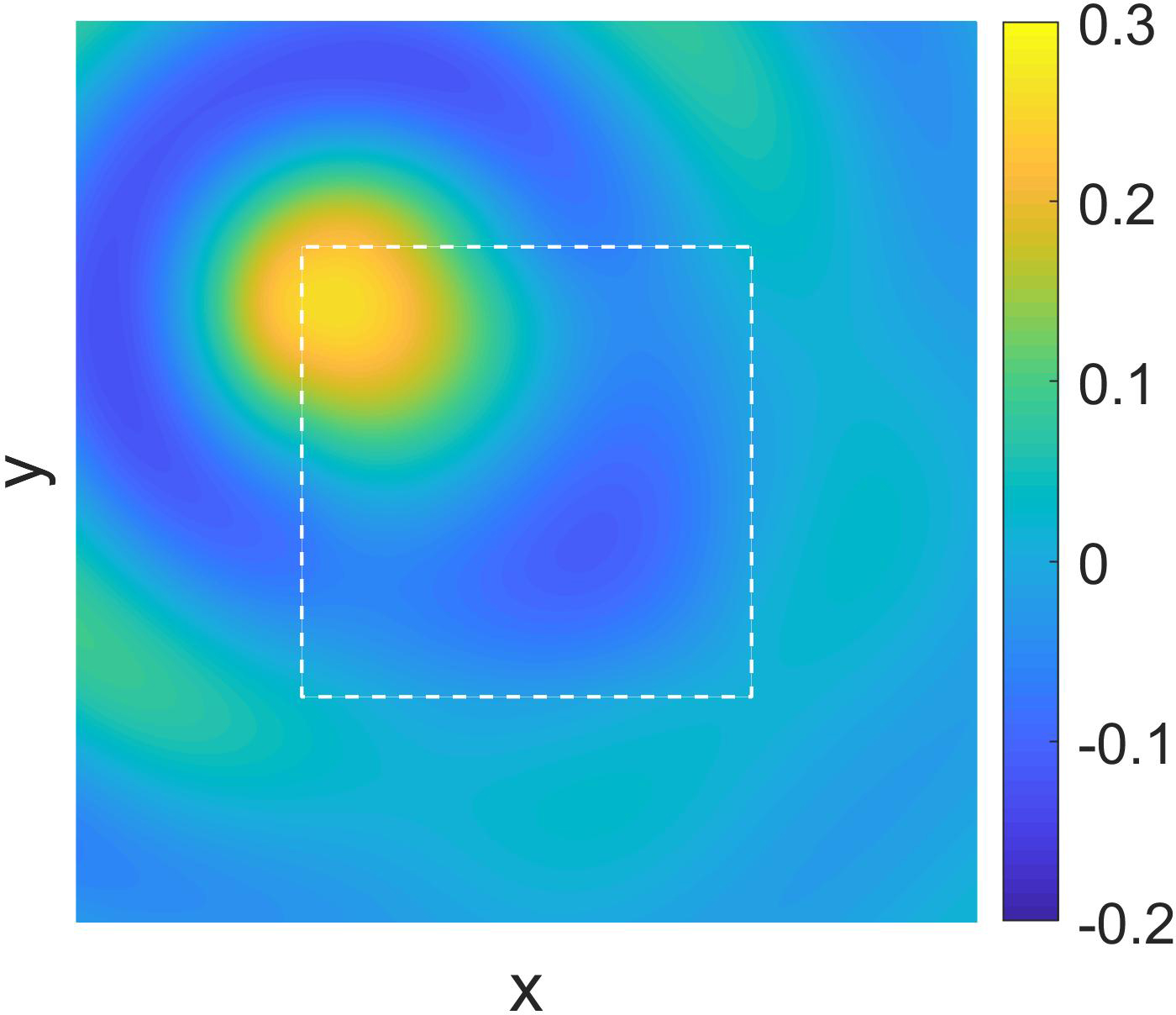}
%\caption{fig2}
}%
\subfigure[Error]{
\centering
\includegraphics[width=0.19\textwidth]{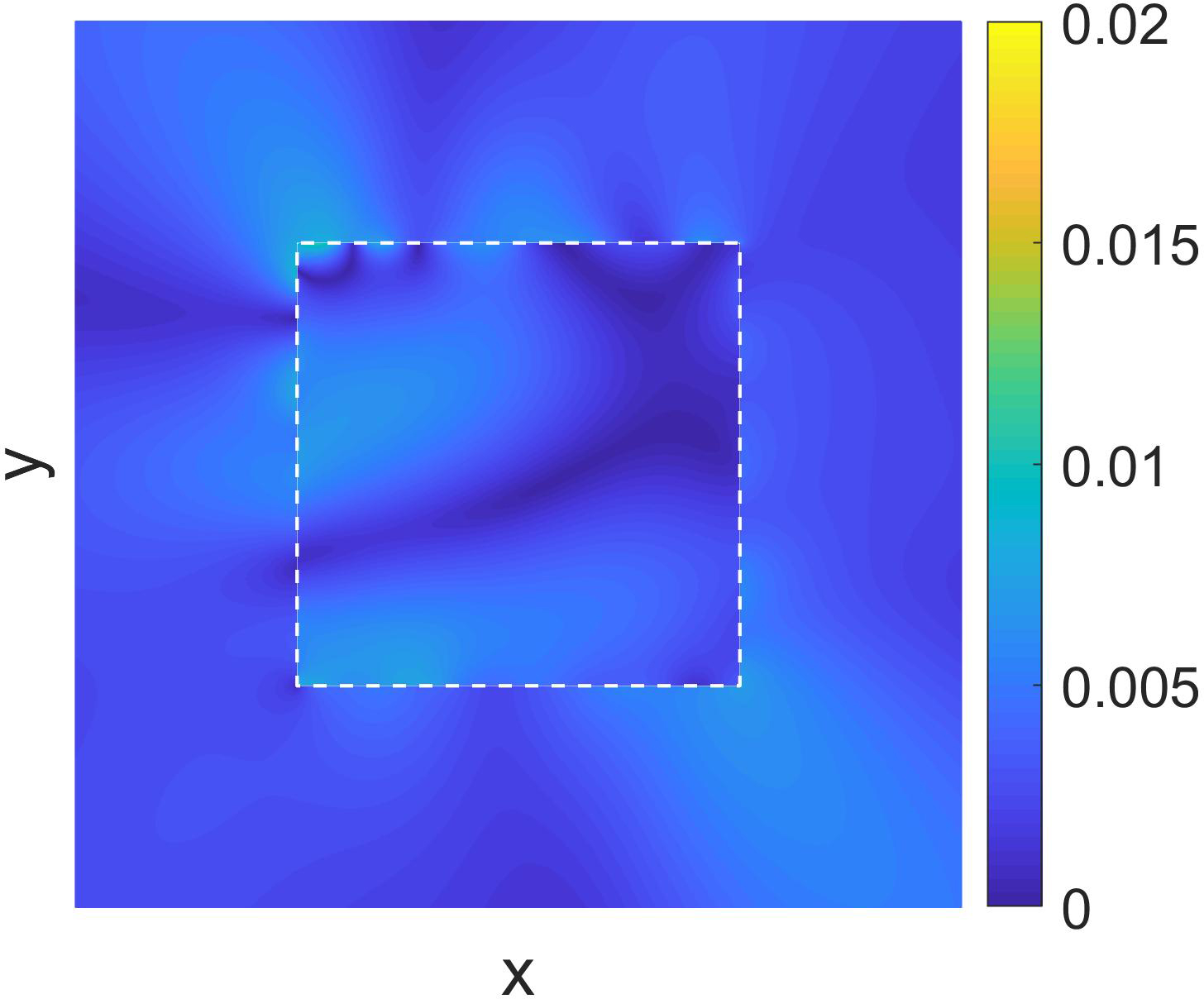}
%\caption{fig2}
}%

\subfigure[Exact, real]{
\centering
\includegraphics[width=0.19\textwidth]{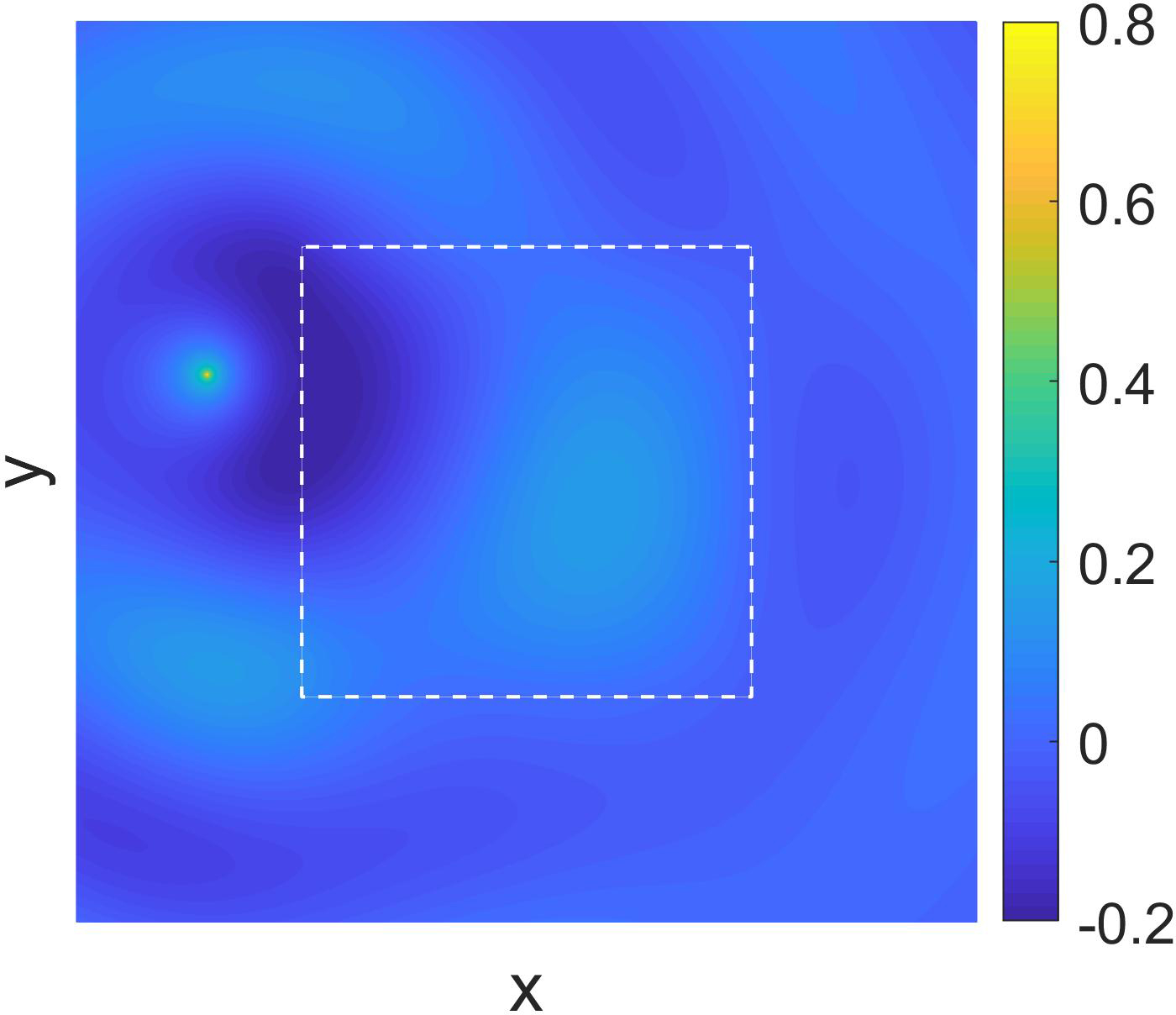}
%\caption{fig1}
}%
\subfigure[Num, real]{
\centering
\includegraphics[width=0.19\textwidth]{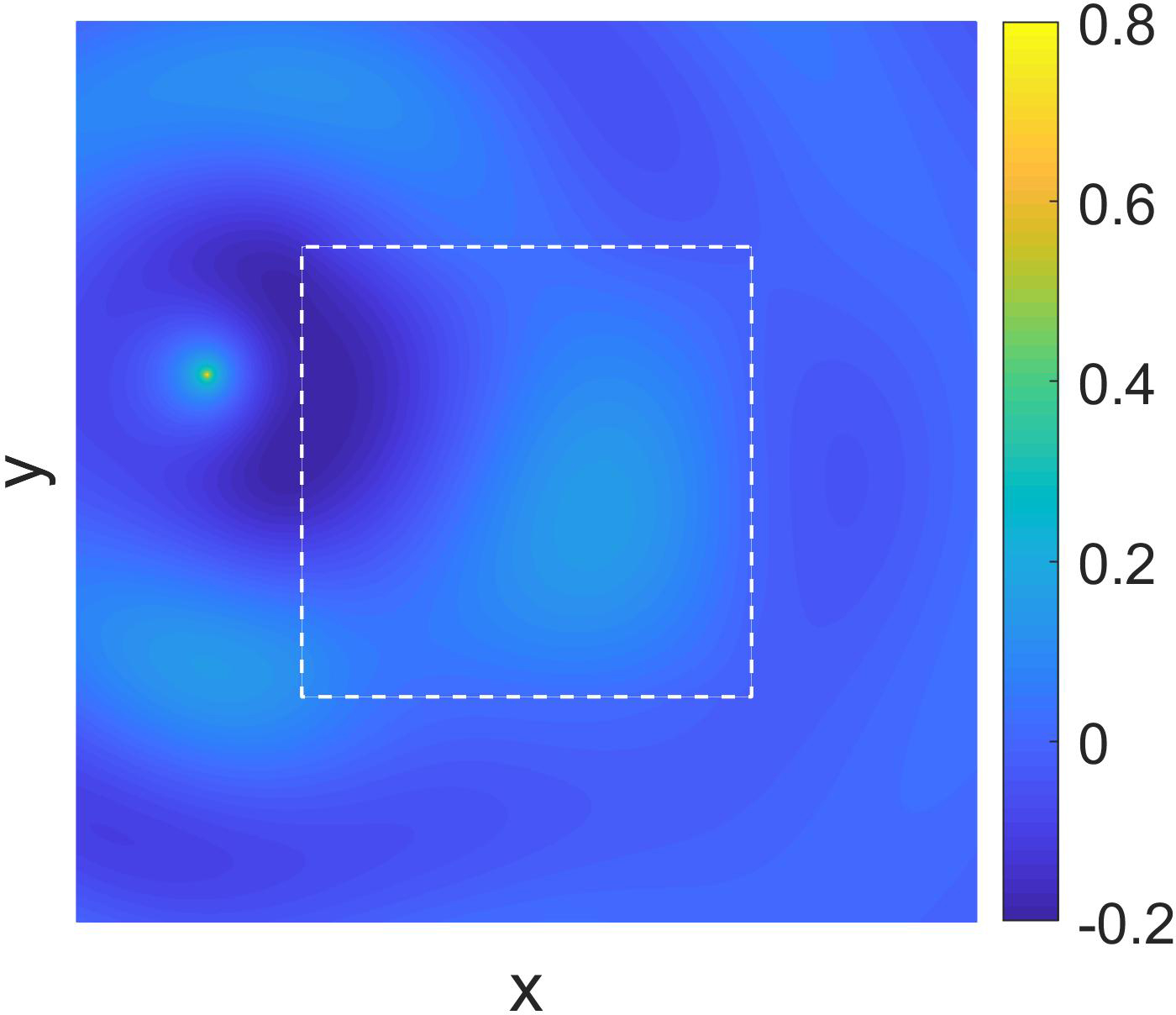}
%\caption{fig2}
}%
\subfigure[Exact, imag]{
\centering
\includegraphics[width=0.19\textwidth]{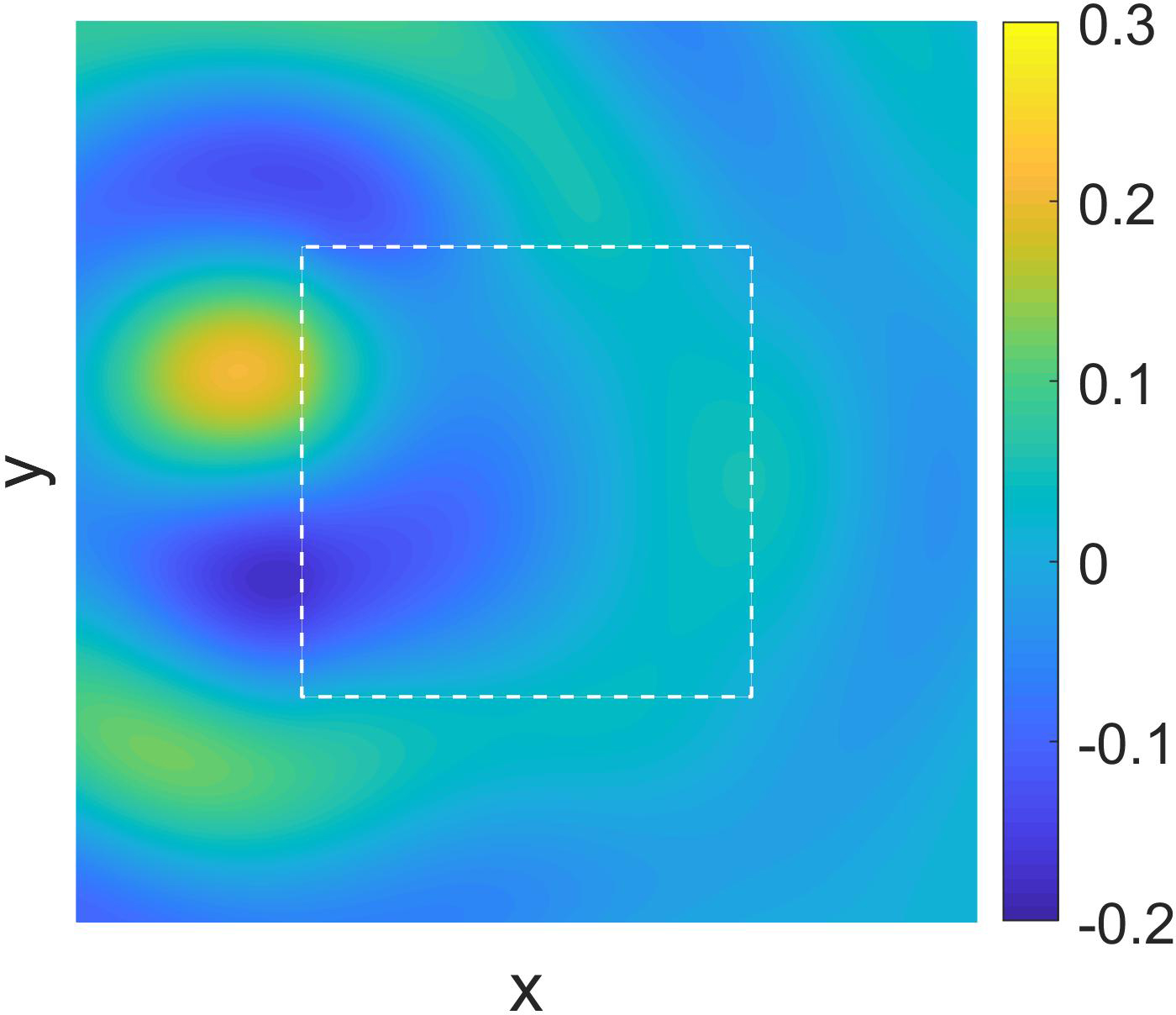}
%\caption{fig2}
}%
\subfigure[Num, imag]{
\centering
\includegraphics[width=0.19\textwidth]{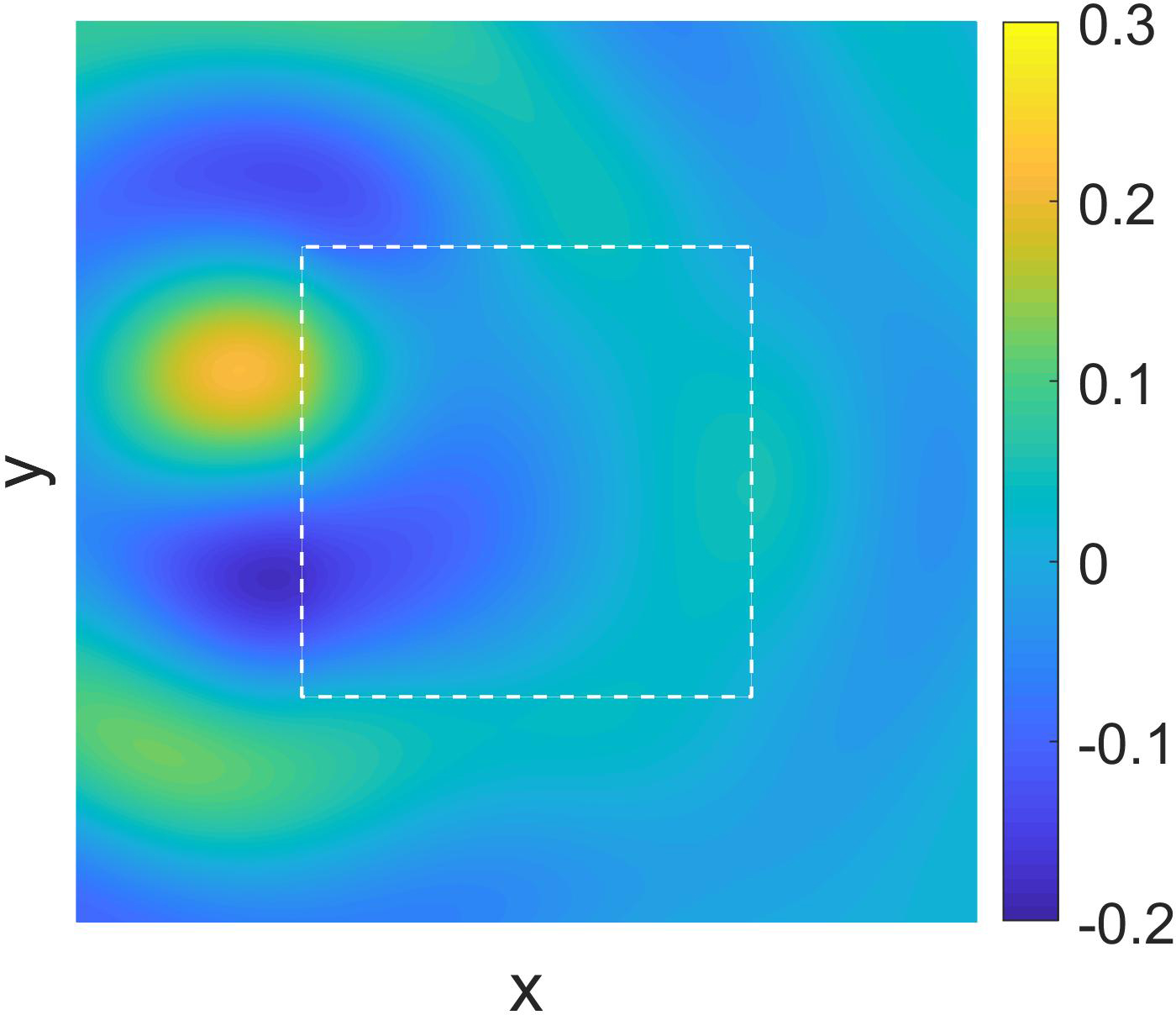}
%\caption{fig2}
}%
\subfigure[Error]{
\centering
\includegraphics[width=0.19\textwidth]{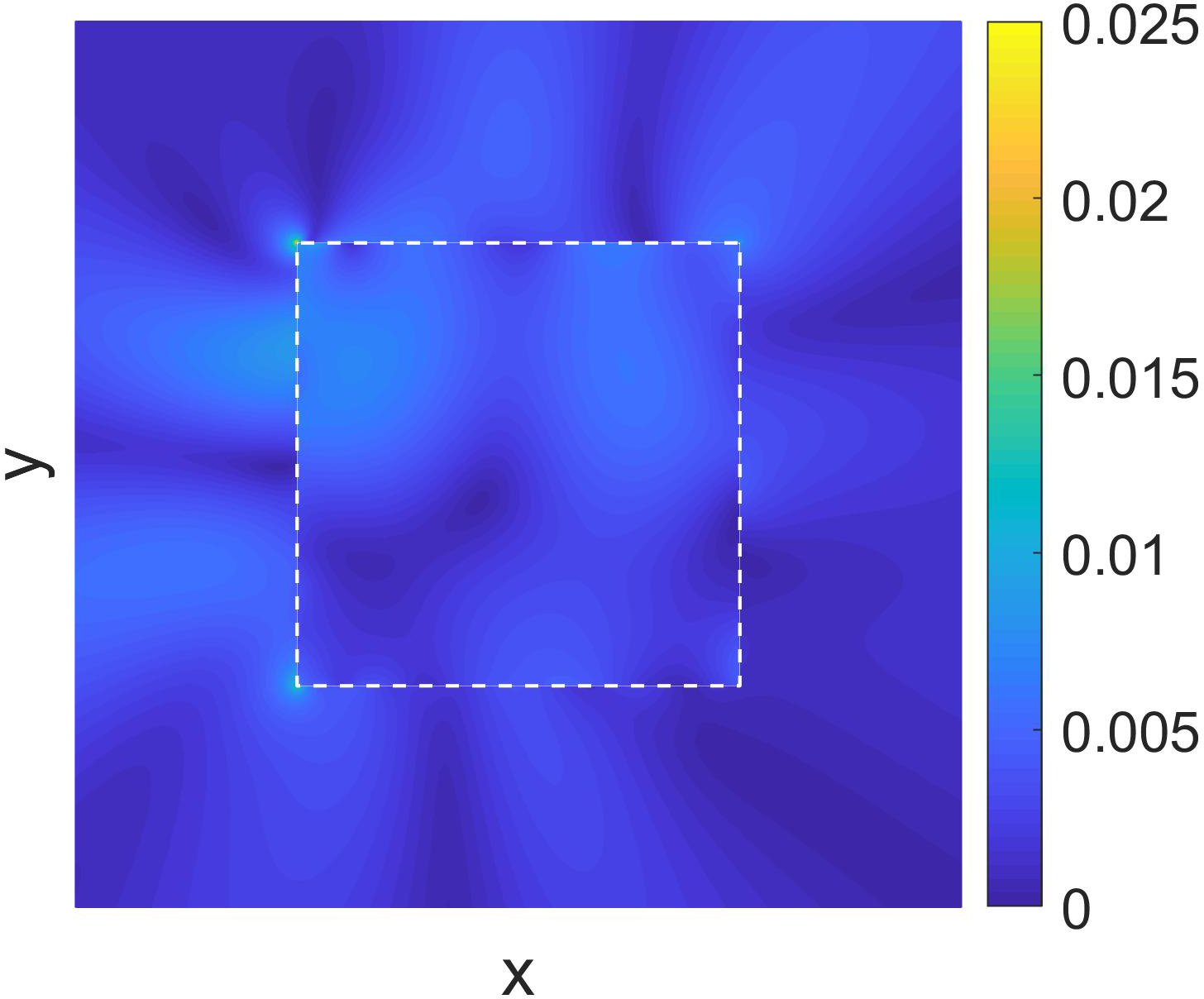}
%\caption{fig2}
}%
\caption{Figures (a)-(e) are for $x=(-0.8559,-0.8164)\in\Omega_1$ with relative $L^2$ error $3.98\%$, while figure (f)-(j) are for $x=(-1.4218,0.4312)\in\Omega_2$ with relative $L^2$ error $2.73\%$. First and second column: the real part of the exact solution and the numerical solution; Third and Forth column: imaginary part of the exact solution and the numerical solution; Fifth column: absolute error between exact solution and numerical solution.}
\label{fig-ex5-1}
\end{figure}

The network and training hyperparameters are set the same as those in the Poisson interface problem. The average relative $L_2$ error of the Green's function $G(x,y)$ is $4.63\times10^{-2}$. For two fixed $x$, the exact Green's function (obtained by traditional boundary integral method), numerical solution obtained by neural network and error is shown in Fig. \ref{fig-ex5-1}, implying the effectiveness of the proposed method in computing Green's function.

Furthermore, after the Green's function is learnt by the neural network, to show the generalization ability of the proposed method in solving PDEs, we consider the homogenuous case of problem (\ref{interface}), i.e., $f\equiv 0$, while the two jump conditions $g_1$ and $g_2$ are generated by the superposition of a class of parameterized function. Specifically, the exact solution to the interface problem is designed as
\begin{equation}
u(x,y)=\left\{
\begin{aligned}
&\sum_{i=1}^I c_1(i)e^{j(k_1(i)x+k_2(i)y)\sqrt{\varepsilon_1\mu_1}},& \quad (x,y)\in\Omega_1,\\
&\sum_{i=1}^I c_2(i)H_0^1(k\sqrt{\varepsilon_2\mu_2}\sqrt{(x-x_0(i))^2+(y-y_0(i))^2}),& \quad (x,y)\in\Omega_2,\\
\end{aligned}
\right.
\end{equation}
where $\{c_1,c_2,k_1,k_2,x_0,y_0\}$ is a set of randomly generated parameters satisfying $k_1^2+k_2^2=k^2$ and $(x_0(i),y_0(i))\in\Omega_1, \ \forall i$. We set $c_1,c_2\sim U[0,1]$, $k_1=k\cos\theta, k_2=k\sin\theta$ with $\theta\sim U[0,2\pi]$, $x_0,y_0\sim U[-0.8,0.8]$. $g_1$ and $g_2$ can be directly computed using the exact solution. Once the Green's function to this interface problem is obtained, the solution to the PDE can be directly computed by
$$u(x)=\int_{\Gamma}\left(\frac{1}{\mu}\frac{\partial G(x,y)}{\partial n_y}g_1(y)-G(x,y)g_2(y)\right)ds_y.$$

\begin{figure}[t]
\centering
\subfigure[Error distribution]{
\centering
\includegraphics[width=0.3\textwidth]{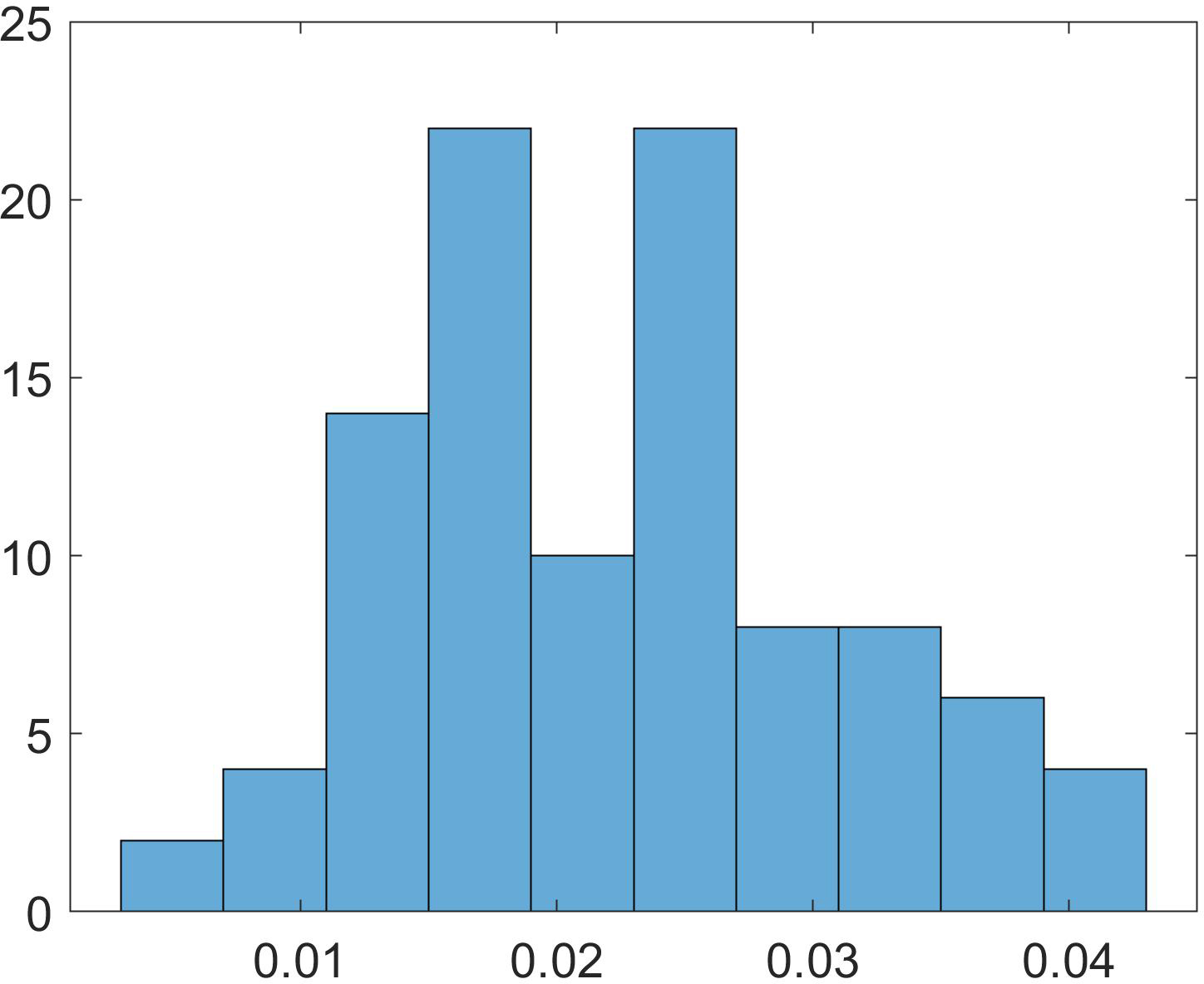}
%\caption{fig1}
}%
\hspace{0.05in}
\subfigure[Exact, real]{
\centering
\includegraphics[width=0.3\textwidth]{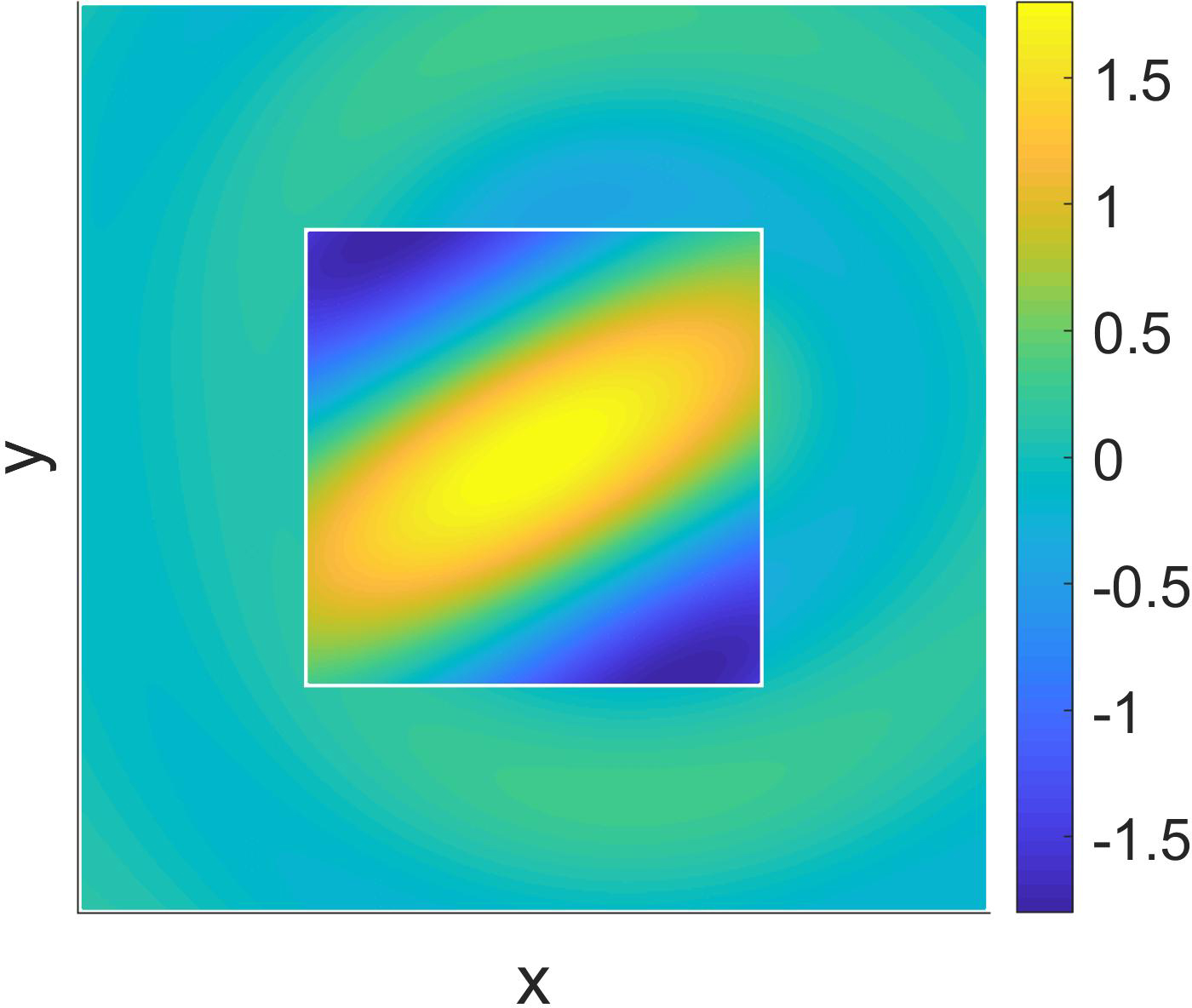}
%\caption{fig1}
}%
\hspace{0.05in}
\subfigure[Numerical, real]{
\centering
\includegraphics[width=0.3\textwidth]{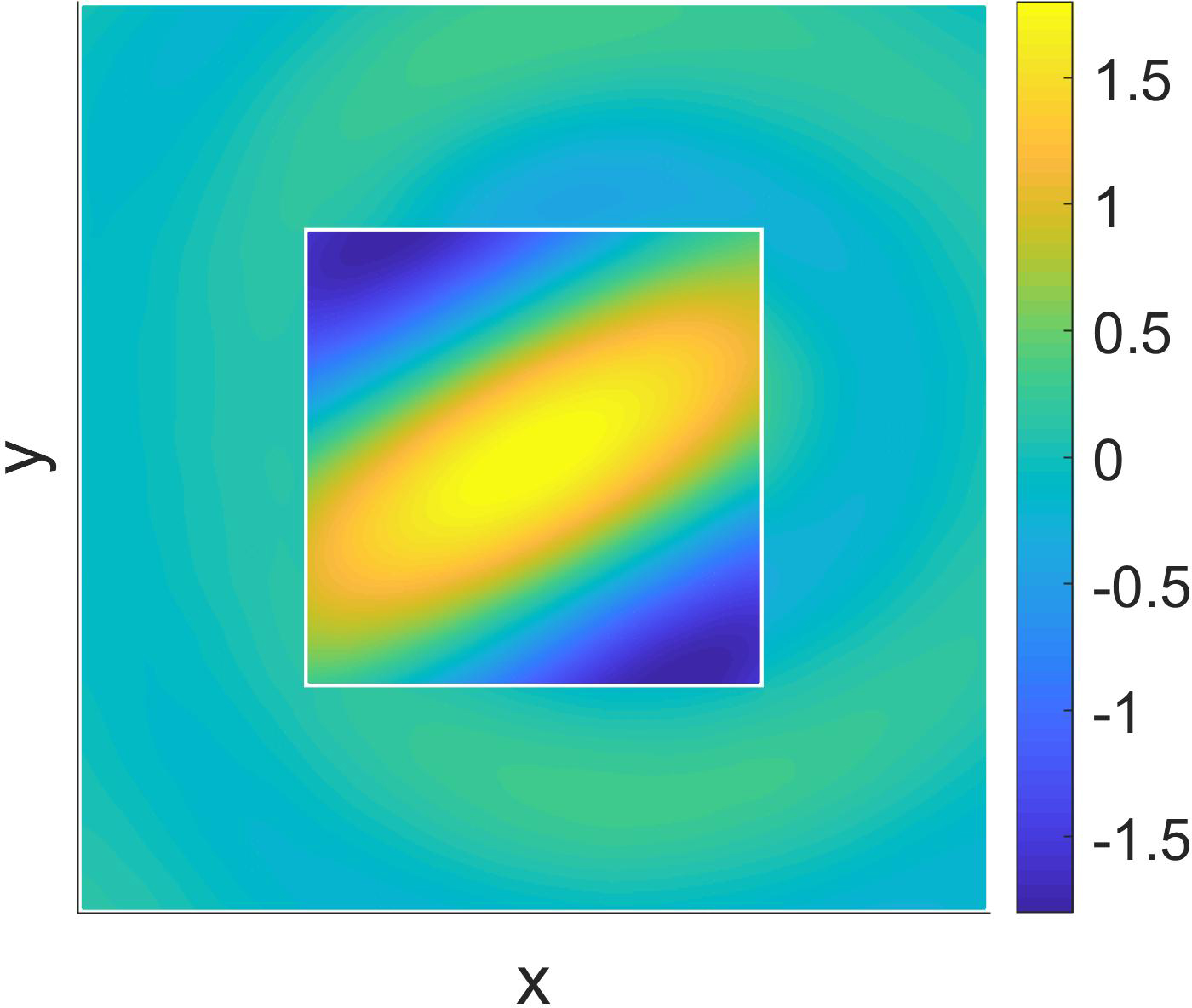}
%\caption{fig2}
}%
\hspace{1in}
\subfigure[Exact, imaginary]{
\centering
\includegraphics[width=0.3\textwidth]{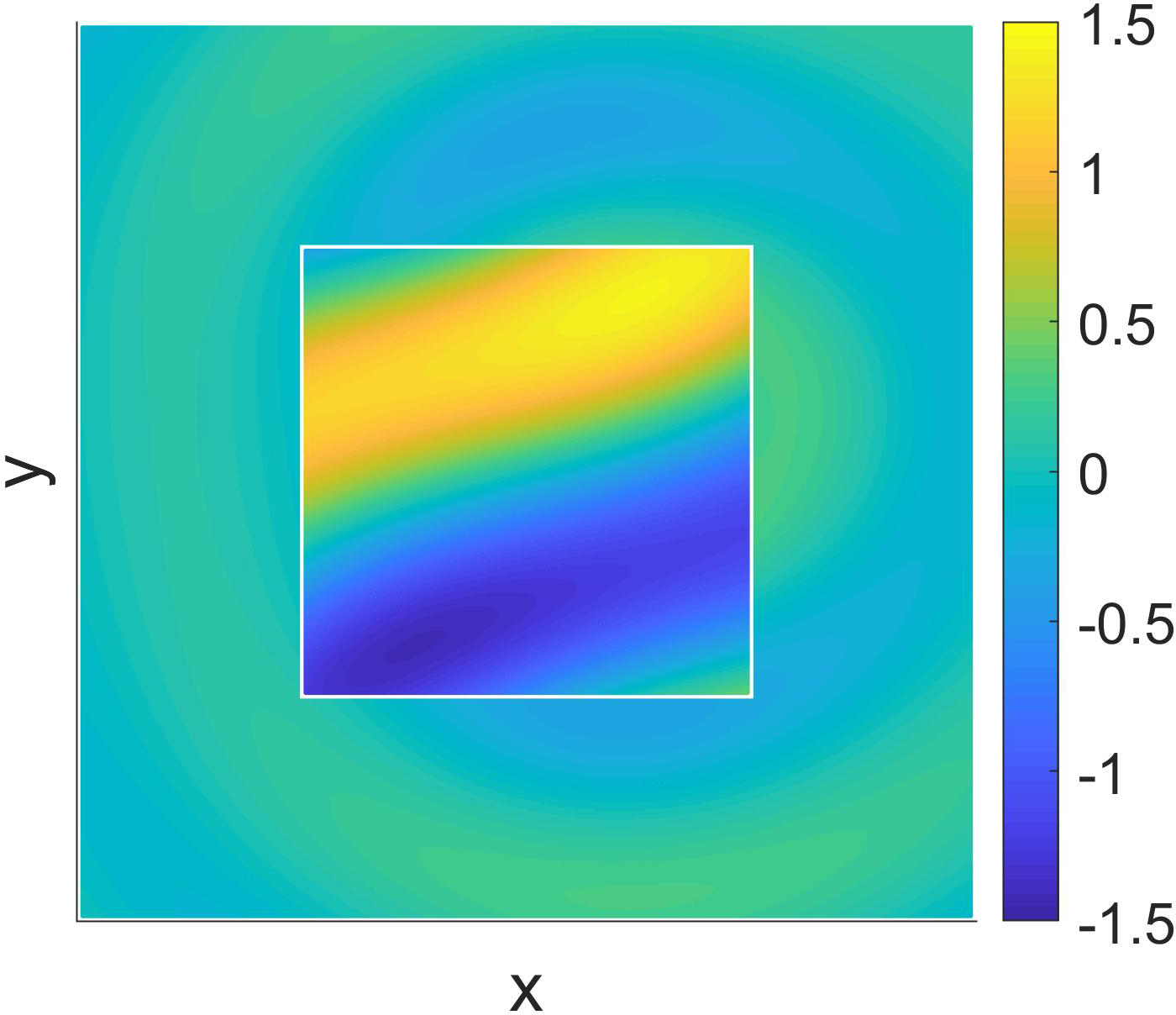}
%\caption{fig2}
}%
\hspace{0.05in}
\subfigure[Numerical, imaginary]{
\centering
\includegraphics[width=0.3\textwidth]{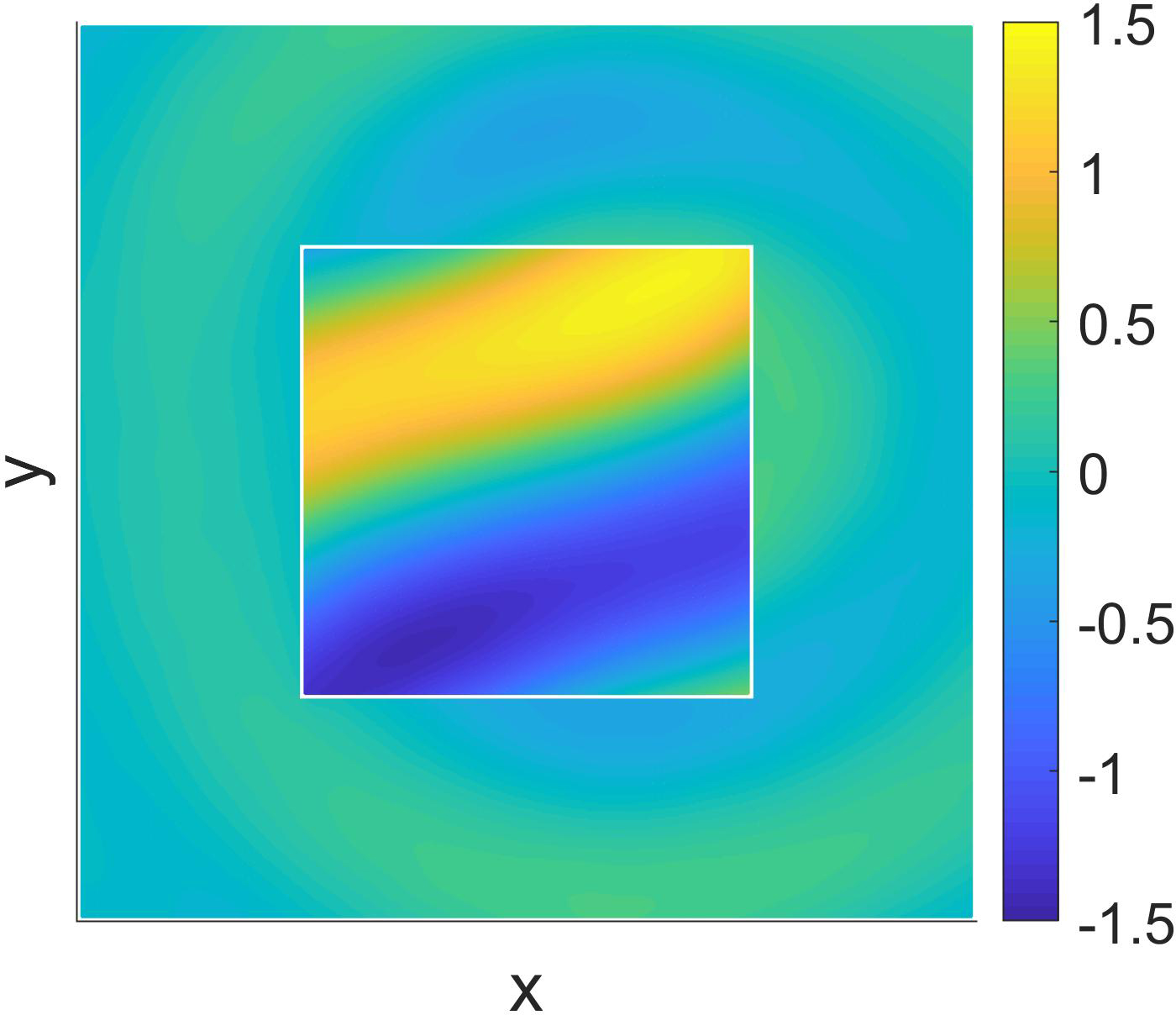}
%\caption{fig2}
}%
\hspace{0.05in}
\subfigure[Error]{
\centering
\includegraphics[width=0.3\textwidth]{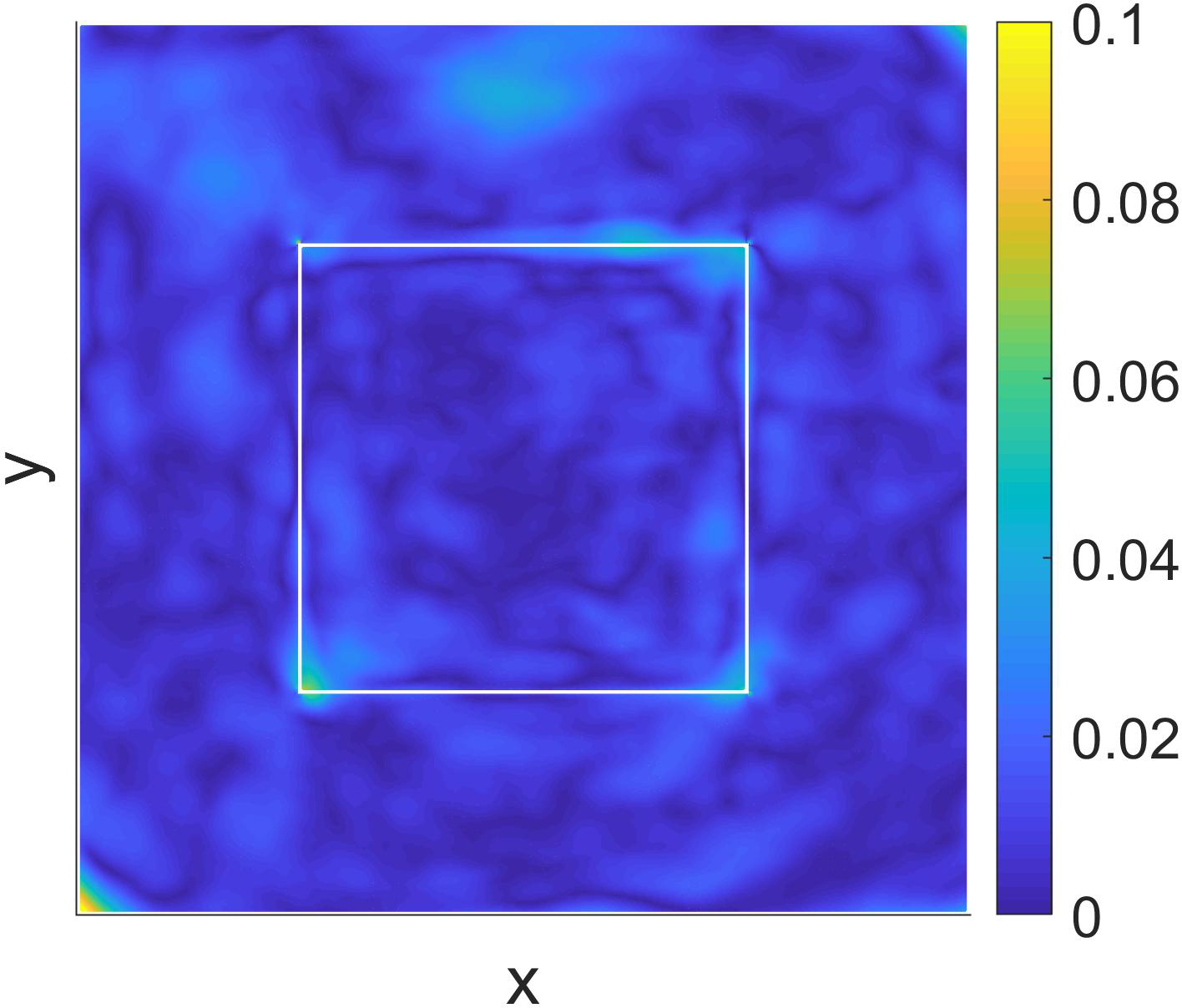}
%\caption{fig2}
}%
\caption{(a): histogram of the relative $L^2$ error of the 100 randomly generated equations. Figures (b)-(f) are the exact solution and numerical solution of one specific equation generated. (b) and (c): real part of the exact solution and the numerical solution; (d) and (e): imaginary part of the exact solution and the numerical solution; (f): absolute error between exact solution and numerical solution.}
\label{fig-ex5-2}
\end{figure}

We take $I=3$ and randomly generate 100 sets of parameters, and solve the interface problem using the learnt Green's function. The histogram of the relative $L^2$ errors of the numerical solutions to the 100 equations are shown in Fig. \ref{fig-ex5-2} (a), while the solution and corresponding error for one set of parameters are given in Fig. \ref{fig-ex5-2}. It can be seen that all the relative $L^2$ errors are below $4\%$. Therefore, not only can the proposed method accurately solve a class of PDEs accurately, but also this method has natural generalization ability over the PDE information. The parameters of the exact solution corresponding to Fig. (\ref{fig-ex5-2}) (b)-(f) are given in Table \ref{table_hel}.
\begin{table}[htbp]
\centering
\renewcommand\arraystretch{1.5}
\caption{Parameter setting for Fig. (\ref{fig-ex5-2}) (b)-(f)}
\vspace{0.1cm}
\label{table_hel}
\begin{tabular}{llll}
\hline
Index & $i=1$  & $i=2$  & $i=3$   \\
\hline
$c_1$ & 0.5550  & 0.9934  & 0.2986  \\
$k_1$ & 1.9959  & 1.6667  & -0.9091 \\
$k_2$ & -0.1282 & -1.1056 & 1.7814  \\
$c_2$ & 0.3963  & 0.3051  & 0.3642  \\
$x_0$ & 0.2850  & 0.1625  & 0.2190  \\
$y_0$ & 0.5724  & -0.1154 & 0.6809  \\
\hline
\end{tabular}
\end{table}

\section{Conclusion}\label{sec:con}
In this paper, a novel neural network based method for learning Green's function is proposed. By utilizing the fundamental solution to remove the singularity in Green's function, the PDEs required for Green's function is reformulated into a smooth high-dimensional problem. Two neural network based methods, DB-GreenNet and BI-GreenNet are propsed to solve this high-dimensional problem. The DB-GreenNet uses the neural network to directly approximate Green's function and take the residual of the differential equation and the boundary conditions as the loss. The BI-GreenNet is based on the 
recently proposed BINet, in which the solution is written in an boundary integral form such that the PDE is automatically satisfied and only boundary terms need to be fitted.

Extensive experiments are conducted and three conclusions can be drawn from the results. Firstly, the proposed method can effectively learn the Green's function of Poisson and Helmholtz equations in bounded domains, unbounded domains and domains with interfaces with high accuracy. Secondly, BI-GreenNet method outperforms DB-GreenNet method in the accuracy of Green's function and the capability to handle problems in unbounded domains. Lastly, the Green's function obtained can be utilized to solve a class of PDEs accurately, and shows good generalization ability over the PDE data, including the source term and boundary conditions.

Although the proposed method exhibits great performance in computing Green's function, the dependence on the fundamental solution hinders the application of this method in varying coefficient problems or equations without explicit fundamental solution. This will be further investigated in the future.

%\backmatter

\bmhead{Acknowledgments}
This work received support by the NSFC under grant 12071244 and the NSFC under grant 11871300.
%Acknowledgments are not compulsory. Where included they should be brief. Grant or contribution numbers may be acknowledged.

%Please refer to Journal-level guidance for any specific requirements.

\section*{Declarations} 
No conflict of interest exits in the submission of this manuscript, and manuscript is approved by all authors for publication. We would like to declare that the work described was original research that has not been published previously, and not under consideration for publication elsewhere, in whole or in part. All the authors listed have approved the manuscript that is enclosed.

%\bibliographystyle{unsrt}
\bibliography{sn-bibliography}% common bib file


\end{document}